\documentclass[9pt,twocolumn,twoside]{pnas-new}
\pdfoutput=1
\templatetype{pnasresearcharticle}

\title{MuA-Ori: Multimodal Actuated Origami}

\author[a,b,c,*]{Antonio Elia Forte}
\author[a,*]{David Melancon}
\author[a,d]{Leon M. Kamp}
\author[a]{Benjamin Gorissen}
\author[a,$\dagger$]{Katia Bertoldi}

\affil[a]{J.A. Paulson School of Engineering and Applied Sciences, Harvard University, Cambridge, MA 02138, USA}
\affil[b]{Department of Electronics, Information and Bioengineering, Politecnico di Milano, Milan, 20133 Italy}
\affil[c]{Department of Engineering, King’s College London, London, WC2R 2LS, UK}
\affil[d]{Department of Applied Physics, Eindhoven University of Technology, Postbus 513, 5600 MB, Eindhoven, The Netherlands}

\leadauthor{Forte} 

\keywords{Origami | Bistabiliy | Inflatables | Actuators} 

\begin{abstract}

Recently, inflatable elements integrated in robotics systems have enabled complex motions as a result of simple inputs. However, these fluidic actuators typically exhibit unimodal deformation upon inflation. Here, we present a new design concept for modular, fluidic actuators that can switch between deformation modes as a response to an input threshold. 
Our system comprises bistable origami modules in which snapping breaks rotational symmetry, giving access to a bending deformation. By tuning geometry, the modules can be designed to snap at different pressure thresholds, rotate clockwise or counterclockwise when actuated, and bend in different planes. Due to their ability to assume multiple deformation modes as response to a single pressure input we call our system MuA-Ori, or Multimodal Actuated Origami. MuA-Ori provides an ideal platform to design actuators that can switch between different configurations, reach multiple, pre-defined targets in space, and move along complex trajectories.
\end{abstract}

\begin{document}

\maketitle
\thispagestyle{firststyle}
\ifthenelse{\boolean{shortarticle}}{\ifthenelse{\boolean{singlecolumn}}{\abscontentformatted}{\abscontent}}{}

\section*{Introduction}

Deformable and inflatable components have increasingly been integrated into robotic systems, as they
provide complex deformations  \cite{Rus2015DesignFA,Hawkeseaan3028}, the ability to morph into target shapes \cite{10.1145/2601097.2601166,Pikul210,konakovi,bico2019,655447}, and inherent compliance, which in turn enables safe interactions \cite{runciman2019soft,polygerinos2015soft,shepherd2019fluidic}.
These, however, suffer from a common drawback:  an intrinsic one-to-one relationship between input pressure and output deformation. Upon inflation in fact, fluidic actuators exhibit unimodal deformation, which accentuates as pressure increases \cite{mosadegh2014pneumatic}.
In order to overcome this and create complex functionality,
multiple actuators can be assembled and carefully sequenced  \cite{multigait,gorissen2019hardware,vasios2019harnessing,Robertsoneaan6357,Bootheaat1853},
or multiple chambers of a single actuator can be pressurized independently \cite{doi:10.1089/soro.2016.0023,doi:10.1089/soro.2013.0009}. As an alternative route, bidirectional bending has been achieved by harnessing material inextensibility \cite{doi:10.1089/soro.2020.0017} and non-linearity \cite{doi:10.1163/016918611X574731,ZENTNER20091009}. 
However, a design strategy that enables arbitrary deformation modes with a single pressure input is still absent.

In an effort to design novel robotic systems, engineers and scientists have recently explored origami principles to create machines that are self-foldable \cite{Felton644,rusnatreviews2018,doi:10.1146/annurev-conmatphys-031016-025316}  and realizable in a variety of materials using both planar and 3D fabrication techniques \cite{Hawkes12441,melanconnature,paulinosoftmatter,paulinosmall}. The functionality of such origami robots can be further expanded if the crease pattern  supports a  non-convex energy  landscape, which enables  multiple stable states \cite{SOROUSH2017,article,kresling,cohennatmat2015,Waitukaitis_2015,PhysRevLett.114.185502,PhysRevE.95.013002,Novelino24096,hypar}. For example, self-locking grippers \cite{Faber1386} and energy-absorbing components for drones \cite{Mintcheveaau0275} were designed out of multistable origami sheets based on the tiling of the degree-four vertex;   bistable mechanical bits and logic elements were created by introducing multistability in the classic waterbomb origami pattern \cite{article,Treml6916,sadeghi2020dynamic} and, finally, bistable configurations of the Kresling pattern \cite{Kresling2,kresling} were exploited to  generate locomotion via peristaltic motion \cite{BHOVAD2019100552} or differential friction \cite{Pagano_2017}, create flexible joints for robotic manipulation \cite{yi_robot_arm}, and store mechanical memory \cite{Novelino24096,natureTachi}. All together, these examples  show that multistable origami are a promising platform to realize  fluidic actuators capable of supporting arbitrary deformation modes when actuated with a single input.

Inspired by the potential of the Kresling pattern in the design of robotic systems, here we employ this classic origami fold as a building block to realize multi-output, but single-input actuators. As part of our strategy, we start with a monostable Kresling pattern and make it bistable by introducing two additional valley creases in one of its panels (see \mbox{Figs.~\ref{fig:fig1}a-b}). During inflation, this panel unfolds and snaps outward, breaking the rotational symmetry of the module. Importantly, upon vacuum such asymmetry gives rise to bending in the opposite direction to the bistable panel up until a critical negative pressure is reached: at this point the panel snaps inward, resetting the module to its initial state. We use a combinatorial approach to couple multiple modules and create actuators with prescribed and complex deformation states that can be reached by controlling a single pressure input. Given that they output multiple deformation modes with a single input, we call our devices Multi-modal Actuated Origami, or MuA-Ori. MuA-Ori offers new opportunities for the design of robotic systems capable of performing complex tasks despite the simple actuation scheme, as demonstrated by the design of a land rowing robot.
\section*{The MuA-Ori building blocks based on  Kresling pattern}

\begin{figure*}[h!]
\centering
\includegraphics[width=1\linewidth]{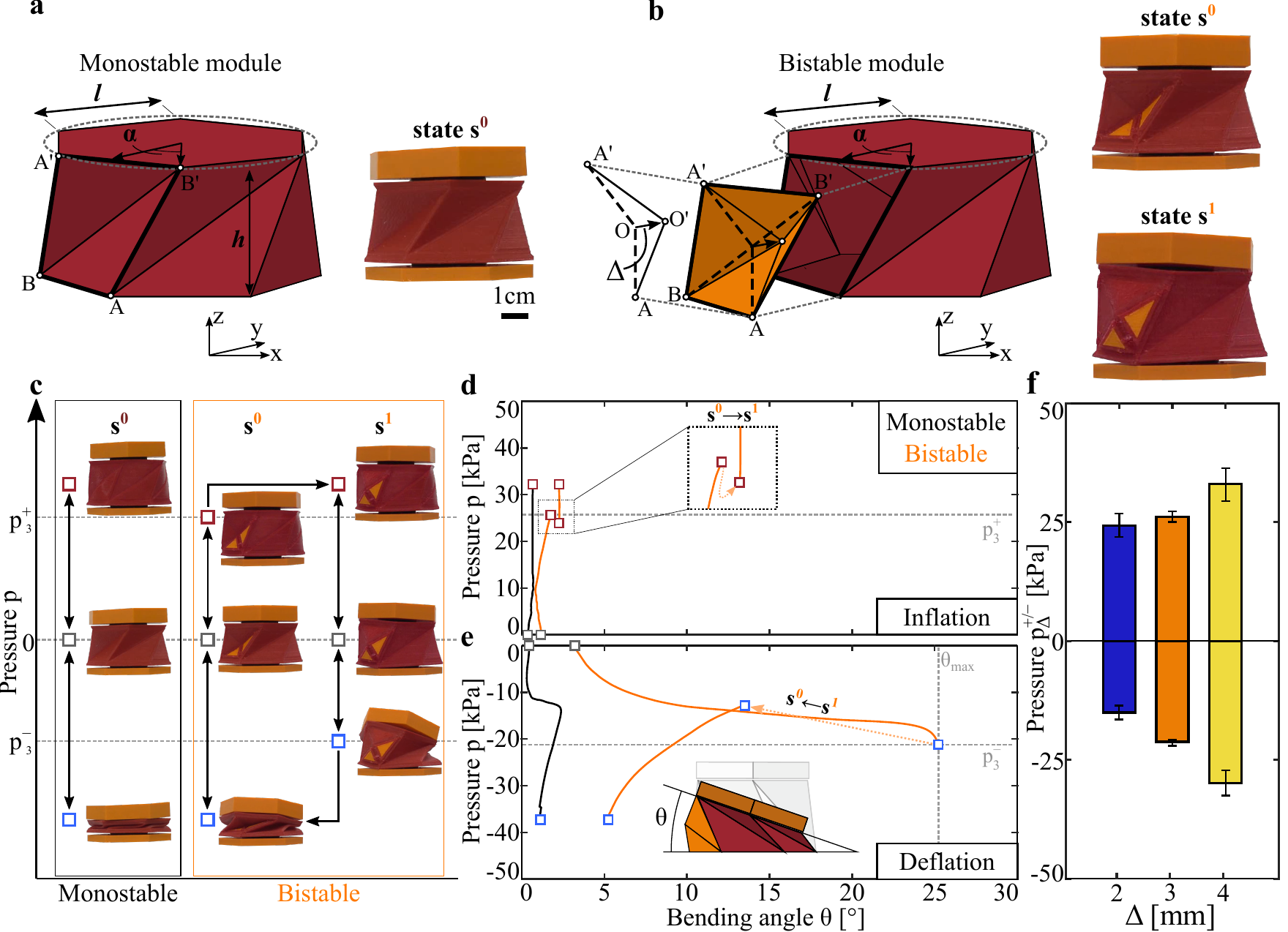}
\caption{\textbf{Bistable origami modules as building blocks for multi-output, single input inflatable actuators}. \textbf{(a)} Schematics of a monostable module based on the hexagonal-base Kresling origami pattern, along with a 3D-printed prototype. The panels of the monostable modules remain always folded inward. We refer to this state of deformation as state $s^0$. \textbf{(b)} {Bistable module with a modified panel (highlighted in orange) made of four triangular facets $A^\prime B O^\prime$, $A O^\prime B^\prime$, $A O^\prime B$, and $A^\prime B^\prime O^\prime$ and characterized by a depth $\Delta$ from vertex $O$ to $O^\prime$, along with a 3D-printed prototype displayed in its two stable states: state $s^0$ for which all panels (including the modified panel) are folded inward; and state $s^1$ for which  the modified panel is popped outward (while all other panels are still folded inward)}. \textbf{(c)} State diagram of the pressurized origami modules. \textbf{(d-e)} Pressure vs.\ bending angle curves for the {monostable} and bistable origami modules during inflation and deflation. At the onset of deflation for the bistable unit, $\theta$ is slightly larger than in the {monostable} module due to the geometrical incompatibility that arises in the structure when the modified panel is in the outward configuration. \textbf{(f)} Experimental positive and negative pressure thresholds, $p^+_\Delta$ and $p^-_\Delta$, as a function of the {modified} panel's depth, $\Delta$.} 
\label{fig:fig1}
\end{figure*}

To realize our nonlinear and reconfigurable MuA-Ori, we use origami building blocks comprising one layer of the classic Kresling pattern (also known as nejiri ori) \cite{Kresling2}. More specifically, {in its initial, undeformed state,} the single module consists of two hexagonal caps with edges of length \mbox{$l = 30 $ mm}, separated by a distance \mbox{$h=24$ mm}, and rotated by an angle \mbox{$\alpha=30^\circ$} with respect to each other (see Fig.~\ref{fig:fig1}a). The hexagons are connected at each side by a panel comprising a pair of triangular facets coupled by alternating mountain (i.e. edges $A^\prime B$ and $AB^\prime$) and valley (i.e. edge $BB^\prime$) creases. {Since the Kresling pattern is not rigid foldable~\cite{kresling}, any change in its internal volume will lead to an incompatible configuration}. To accommodate the resulting geometrical frustration, we 3D-print $1$-mm thick triangular facets out of a compliant material (TPU95A from Ultimaker with Young's modulus \mbox{$E = 26$ MPa}) and reduce the thickness locally to \mbox{$0.4$ mm} to create the hinges (see prototype in Fig.~\ref{fig:fig1}a). Further, to facilitate coupling between different modules, we 3D-print the hexagonal caps out of a stiffer material ({PLA from Ultimaker with Young's modulus \mbox{$E = 2.3$ GPa}}) {and,  to form an inflatable cavity, we coat the origami unit with a thin layer of polydimethylsiloxane (PDMS) (see Supplementary Materials, Section \textcolor{blue}{S1} for fabrication details)}. {Finally, it is worth noticing that the chosen values of the parameters $(h,l,\alpha)$ yield a monostable origami module (i.e. the Kresling pattern is only stable in its initial, undeformed state).}

To investigate the response of a single module, we inflate it with water while submerged in water (to eliminate the effects of gravity - see Supplementary Materials, Section \textcolor{blue}{S2} for details). {As expected \cite{kresling}, the Kresling unit extends and contracts (while twisting at the same time) upon inflation and deflation (Fig.~\ref{fig:fig1}c) and returns to its undeformed state as soon as the pressure is removed. This state of deformation, in which all panels are folded inward, is named $s^0$. When in this state the bending angle, $\theta$, remains close to zero (Figs.~\ref{fig:fig1}d-e).} 

Aiming at unlocking different deformation modes with one single pressure input, we then take inspiration from bistability in degree-four vertices \cite{Waitukaitis_2015,sadeghi2020dynamic,PhysRevE.94.043002} and modify one of the original Kresling panels by introducing two additional valley creases (i.e. $AO$ and $A^\prime O$ with $O$ being the midpoint of crease $BB^\prime$, see Fig.~\ref{fig:fig1}b). While this effectively creates a degree-four vertex, it results in a monostable origami unit, as no snap-through instability is recorded upon inflation (see Supplementary Materials, Section \textcolor{blue}{S2} for details). To increase the geometric incompatibility during deployment and achieve bistability in the unit, we then move the degree-four vertex inward by $\Delta$ (see Fig.~\ref{fig:fig1}b where $\Delta$ is the norm of vector $\overline{OO^\prime}$ perpendicular to vectors $\overline{AA^\prime}$ and $\overline{BB^\prime}$). Choosing \mbox{$\Delta = 3$ mm}, for example, we can fabricate an origami unit that can easily transition between two stable states: {state $s^0$ for which all panels are folded inward, and state $s^1$ for which the modified panel is popped outward (while all other panels are still folded inward).} Similar to the unit based on the classic Kresling pattern, upon inflation this modified module simply extends with all panels bent inward if $p< 26.1\pm0.9$ kPa. However, at \mbox{$p_3^+ = 26.1\pm0.9$ kPa} (where the subscript refers to $\Delta = 3$ mm {and the superscript refers to positive pressure}), the unit snaps from state $s^0$ with the modified panel folded inward to state $s^1$ where it is popped  outward (Fig.~\ref{fig:fig1}c)---a transition which is accompanied by a discontinuity of the bending angle $\theta$ (see zoom-in in Fig.~\ref{fig:fig1}d). Finally, a further increase in pressure causes the unit to elongate until the maximum structural limit is reached. {Afterward, when the input pressure is removed,   the modified panel remains popped outward (see Fig.~\ref{fig:fig1}c) because of bistability. As such, when we apply negative pressure the unit not only contracts, but also bends (see Figs.~\ref{fig:fig1}c and e), exhibiting a behaviour that radically differs from that of the Kresling module.} {Specifically, we find that at first $\theta$ monotonically increases until the two hexagonal caps come into physical contact effectively clipping the available range of bending deformation to $\theta_{max} =21.7\pm0.3\,^\circ$. As previously mentioned, this bending deformation is caused by the modified panel, which remains in the popped outward configuration (while the other panels fold under increasing negative pressure) and breaks the radial symmetry}. Finally, when the negative pressure passes the threshold $p_3^-=-21.2\pm 0.7$ kPa ({where the superscript refers to negative pressure}), the modified panel snaps back to the inward position (see snapping transition from state $s^1$ to state $s^0$ in Fig.~\ref{fig:fig1}c) and $\theta$ suddenly decreases (dashed orange arrow in Fig.~\ref{fig:fig1}e). If one continues to apply negative pressure to the module, the unit folds (almost) flat, as clearly shown by the decreasing trend of $\theta$. 

Next, we investigate the effect of the depth $\Delta$ of our {degree-four vertex panel} on the positive and negative pressure thresholds, $p_\Delta^+$ and $p_\Delta^-$, as well as the deformed configurations reached upon snapping. The experimental results reported in Fig.~\ref{fig:fig1}f for $\Delta=2$, $3$, and $4$ mm indicate that the absolute value of the pressure thresholds increases with $\Delta$ within the considered range. Differently, we find that the  bending angle, the axial displacement of the end caps and their relative twist remain almost constant with $\Delta$. In fact, the first reaches an upper limit when the caps get in contact with each other and the other two are dominated by the geometric characteristics of the Kresling pattern (see Fig.~\textcolor{blue}{S7}). {Note that for \mbox{$\Delta < 2$ mm},  the modules are found to be monostable. This means that negligible bending is recorded upon application of  negative pressure, since the degree-four vertex panel snaps back immediately.  Differently,  for \mbox{$\Delta \ge 4$ mm}, the positive pressure required to snap the modified panel outward is so high that the module fails (see Fig.~\textcolor{blue}{S6}}).
\section*{Extending the design space for complex outputs}

\begin{figure}[b!]
\centering
\includegraphics[width=0.9\linewidth]{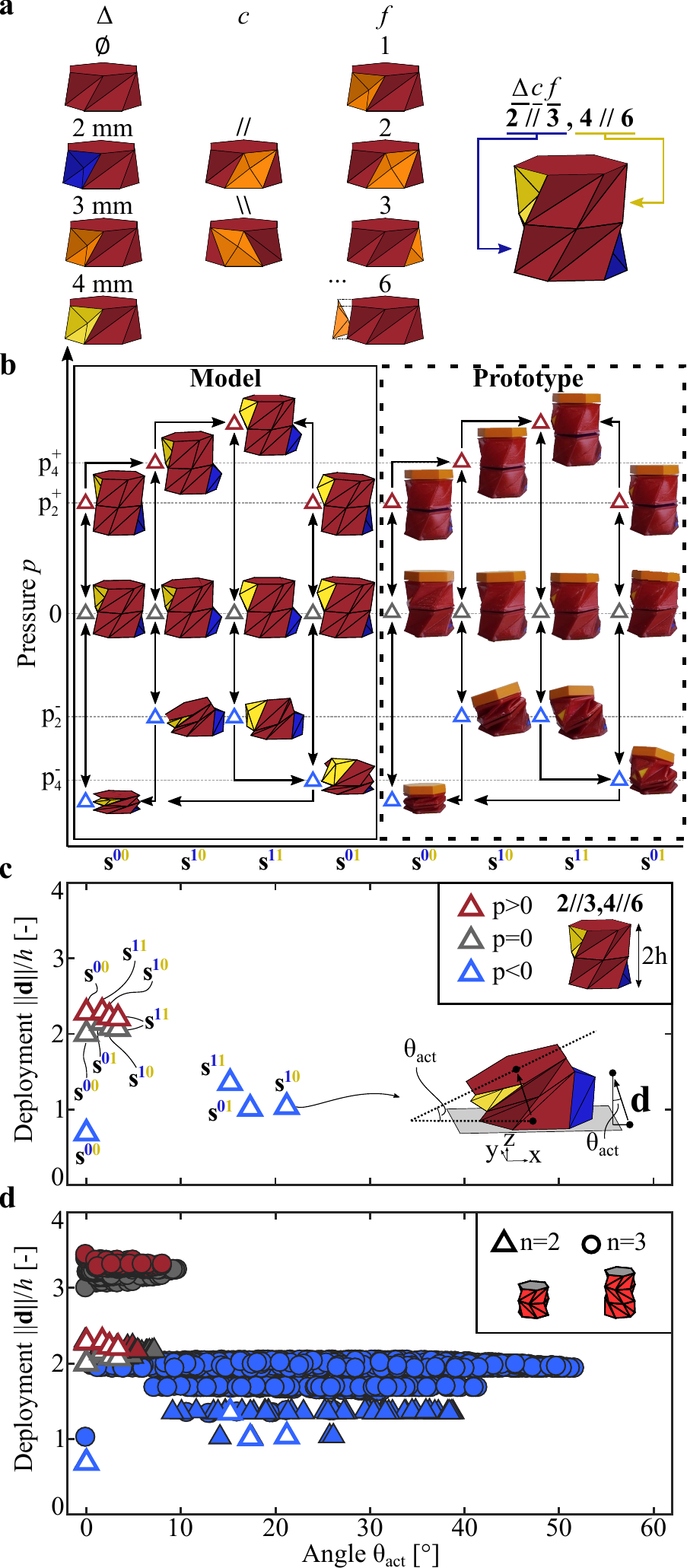}
\caption{\textbf{Extending the design space.} \textbf{(a)} For each $k$ module we define $3$ geometrical parameters:  $\Delta^k$, $c^k$, and $f^k$. Note that, for simplicity, we also assign $f^1$, but only the location of the modified panel relative to its neighbor is important. \textbf{(b)} State diagram for a $2$-units actuator characterized by $[\Delta^1c^1f^1;\Delta^2c^2f^2] = [2\textrm{//}3;4\textrm{//}6]$: geometrical model prediction on the left, experimental snapshots on the right. \textbf{(c)} Normalized deployment $||\mathbf{d}||/h$ and angle $\theta_{act}$ for every stable state of the $2$-units actuator characterized by $[\Delta^1c^1f^1;\Delta^2c^2f^2] = [2\textrm{//}3;4\textrm{//}6]$. \textbf{(d)} Normalized deployment $||\mathbf{d}||/h$ and angle $\theta_{act}$ for every stable state of any possible $2$-units (triangles) and 3-units (circles) actuator in the design space.}
\label{fig:fig2}
\end{figure}

After demonstrating that our bistable module can transition between different stable states with distinct deformation modes (i.e. extension, twisting, and bending), we next combine these units to form actuators able to achieve complex outputs: the MuA-Ori. By combining $n$ modules, we can construct {$\left(3\times2\times6+1\times2\right)^n = 38^n$} different actuators since for each module $k$ we can select ($i$) either a regular Kresling module or a unit comprising a {modified, degree-four vertex} panel with depth $\Delta^k \in \{2,3,4\}$ mm; ($ii$) the upper cap to be rotated clockwise or anticlockwise with respect to the bottom one, $c^k\in \{\textrm{//},\textrm{\textbackslash \textbackslash}\}$,  and ($iii$) the side on which the {modified} panel is located, $f^k\in \{1,\ldots,6\}$ (see Fig. \ref{fig:fig2}a). {Note that the domain of all constraints is discrete since we choose a limited set of values for $\Delta$ and all end plates to have their sides aligned}.

To navigate the vast design space systematically and efficiently, we develop a simple algorithm that predicts all stable states and snapping transitions for actuators comprising $n$ modules upon inflation and deflation. First, we extract key geometric features from the experiments conducted on single units, i.e. height, twisting, and bending angle of the top cap associated to all stable states and snapping transitions (see Figs.~\textcolor{blue}{S7-S8}). When assuming pressure continuity, these data allow the prediction of all configurations for the stable states and snapping transitions of any $n$-units actuator (see Supplementary Materials, Section \textcolor{blue}{S3} for details on the algorithm). Note that we also assume perfect coupling between units, so that the pressure thresholds, $p_{\Delta}^{+/-}$, found in the experimental characterization of Fig.~\ref{fig:fig1}f, remain unchanged and identical for units with the same geometrical parameters. 

As an example, in Fig.~\ref{fig:fig2}a, we consider an actuator comprising  $n=2$ modules characterized by $[\Delta^1c^1f^1;\Delta^2c^2f^2] = [2\textrm{//}3;4\textrm{//}6]$, where we assume the first unit to be the one at the bottom. As expected, this MuA-Ori has four stable states at atmospheric  pressure, $s^{ij}$ (where the subscripts $i,j=0,\,1$, refer to the state of the modified panels, with $\Delta=2$ and 4 mm, respectively), and six snapping transitions that can be triggered by varying the internal pressure (Fig~\ref{fig:fig2}b). Whereas the stable states $s^{10}$ and $s^{11}$ can be readily obtained by simply increasing pressure, a more complex pressure path is required to achieve state $s^{01}$, as one has to ($i$) increase pressure above $p^+_4$ and then ($ii$) decrease it below $p_2^-$. To validate the model, we then conduct experiments on units connected via 3D-printed screws (see Fig.~\textcolor{blue}{S2} for details) and find that they closely match the predictions (see Fig~\ref{fig:fig2}b and Movie \textcolor{blue}{S3}).

Since our final goal is to build actuators with programmed deformation modes, we further quantify the deployment of our actuator by {recording the vector connecting the caps' centroids, $\mathbf{d}$, at each stable state and snapping transition. In Fig.~\ref{fig:fig2}c we report the norm of  $\mathbf{d}$ normalized by the height of a single module, $||\mathbf{d}||/h$, and the angle between each $\mathbf{d}$ and the $z$-axis, $\theta_{act}$, for all $12$ configurations supported by the  actuator characterized by $[\Delta^1c^1f^1;\Delta^2c^2f^2] = [2\textrm{//}3;4\textrm{//}6]$}. We find that in the positive pressure regime the achievable deformation is limited to {simple} extension, resulting in configurations all clustered in a portion of the design space characterized by $\theta_{act} \sim 5^\circ$ and  $||\mathbf{d}||/h\sim 2$ (gray and red triangles). By contrast, behaviors characterized by large bending angles ($\theta_{act} \sim 20^\circ$) unfold under negative pressure (blue triangles).  Importantly, such bending  provides opportunities for the design of actuators that can reach a wider range of configurations. This becomes even clearer when expanding the range of possible configurations by varying the parameters $\Delta^k$, $c^k$, and $f^k$ and considering not only $n=2$ but also $n=3$ building blocks (triangles and circles in Fig.~\ref{fig:fig2}d, respectively). Once again, we notice that all configurations under positive and atmospheric pressure cluster on low values of bending angle and a normalized deployment that is roughly equal to the number of building blocks considered. Differently, in the negative pressure regime a rich set of deformations unfolds, characterized by a wide range of achievable $\theta_{act}$ (the maximum values of $\theta_{act}$ increases from $40^\circ$ to $50^\circ$ from $n=2$ to $n=3$) at near-constant normalized deployment ($||\mathbf{d}||/h$ increases slightly from about $1.5$ to $2$ from $n=2$ to $n=3$). {These results indicate that by stacking our origami modules to form arrays, we can realize actuators capable of supporting a variety of deformation modes, which can be selected by varying the level of applied pressure.}

\section*{Inverse design to reach multiple targets}

{Motivated by the results in  Fig.~\ref{fig:fig2}, we then investigate the behavior of arrays with a larger number of units. Our goal is to build actuators capable of switching between target deformation modes when inflated with a single pressure source. However, since the use of $n$ modules leads to $38^n$ possible actuator designs (i.e. $54,872$ for $n=3$ and $2,085,136$ for $n=4$), it is crucial to use a robust algorithm to efficiently scan the range of responses that can be achieved and identify
\begin{figure}[h!]
\centering
\includegraphics[width=1\linewidth]{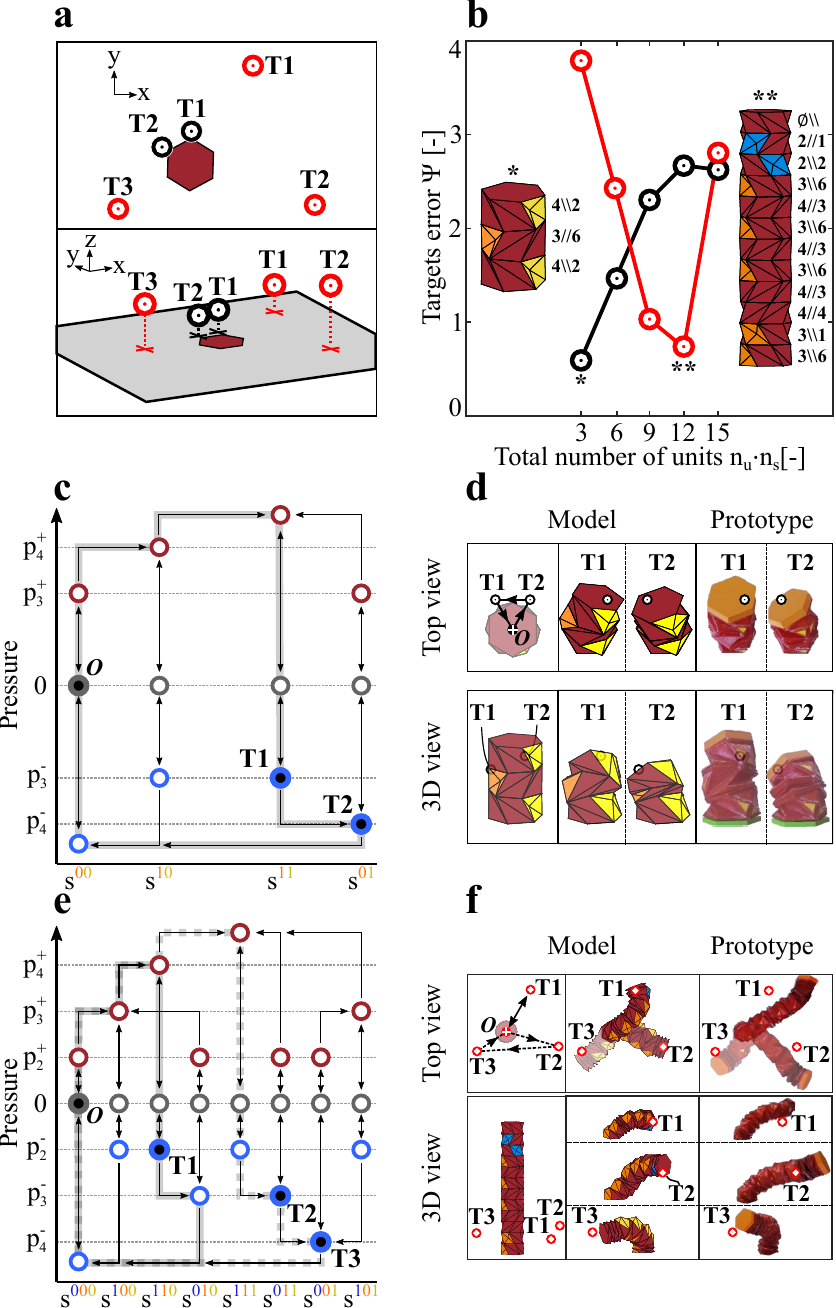}
\caption{\textbf{Inverse design to reach multiple targets}. {We employ the  greedy algorithm to inverse design actuators able to reach a set of targets with a single pressure source}. \textbf{(a)} Two distinct set of targets (black and red), top and 3D view. \textbf{(b)}Targets error $\Psi$ as a function of total number of units: the optimal actuators produced by the algorithm for both set of targets are reported as ($*$) and ($**$), along with the respective parameters for each module. \textbf{(c)} State diagram for the 3-units actuator ($*$) with targets $T1$ and $T2$ highlighted. \textbf{(d)} Top and 3D view of the model and the experimental prototype for the 3-units actuator reaching the black targets. \textbf{(e)}  State diagram for the 12-units actuator ($**$) with targets $T1$, $T2$ and $T3$ highlighted. \textbf{(f)} Top and 3D view of the model and the experimental prototype for the 12-units actuator reaching the red targets.} 
\label{fig:fig3}
\end{figure}
configurations leading to the target deformation modes. Toward this end,  given the discrete nature of our design variables, we use a greedy algorithm based on the best-first search method \cite{CURTIS2003125,bestfirst}---a  progressive local search algorithm that, at each iteration, minimizes the cost function by looking at a set of available solutions. Although there exists many algorithms to solve this type of discrete optimization problems \cite{ga_cite,surrogate}, we find that the greedy algorithm provides the best trade-off between accuracy and computational cost (see Supplementary Information, Section \textcolor{blue}{S4} for details and comparison of the different algorithms). Specifically, our greedy algorithm identifies MuA-Ori comprising $n$ units built out of $n_s$ super-cells each with $n_u$ modules (so that $n=n_u\cdot n_s$) whose tip can reach a desired set of targets arbitrarily positioned in the surrounding space.  At the first iteration, the algorithm starts by selecting the actuator super-cell design  that minimizes
\begin{equation}\label{Psi}
    \Psi = \frac{1}{n_{targets}\cdot h}\sum_{m=1}^{n_{targets}}\min ||\mathbf{d} - \mathbf{T}_m||,
\end{equation}
where $n_{targets}$ is the number of targets, and $\mathbf{T}_m$ is the vector connecting the $m$-th target with the origin. Once the first super-cell is chosen, the algorithm stores it in memory and starts a second iteration. This comes to an end when the algorithm identifies a second super-cell that, connected to the first, one minimizes Eq.~(\ref{Psi}). The first two  super-cells are then stored in memory and the algorithm advances to the next one. Note that in this study, to balance the number of available designs and computational cost, we set the greedy algorithm to consider super-cells made of $3$ units (i.e.~$n_u=3$)  (see Figs.~\textcolor{blue}{S15} and \textcolor{blue}{S18} for a comparison across super-cells made with different $n_u$) and, to avoid fabricating excessively long actuators whose response could be affected by gravity, terminate the algorithm after stacking $5$ super-cells.}

{To demonstrate our approach, we select two distinct sets of targets within the reachable space (see black and red circular markers in Fig.~\ref{fig:fig3}a and Supplementary Materials Figs.~\textcolor{blue}{S15-S19} for additional targets). For the first set of black targets $(T1,T2)$ with small height and radius of action, we find that the objective function $\Psi$ is minimized for an actuator with $n_s = 1$ and  $[\Delta^1c^1f^1;\Delta^2c^2f^2;\Delta^3c^3f^3] = [4\textbackslash \textbackslash2;3//6;4\textbackslash \textbackslash 2]$ (see Fig.~\ref{fig:fig3}b).} As this $3$-unit actuator only comprises modules with $\Delta = 3$ and $4$ mm, its state diagram (shown  in Fig.~\ref{fig:fig3}d) is qualitatively similar to the one reported in Fig.~\ref{fig:fig2}b. To reach the target $T1$, the internal pressure must first be raised above $p^{+}_4$ and then lowered to $p_3^-$ (see shaded gray line in Fig.~\ref{fig:fig3}c). Successively, the target $T2$ is reached by further lowering the pressure to $p_4^-$. It is worth noticing that, in this example, increasing the pressure above $p_4^+$, decreasing it below $p_4-$ and finally bringing it back to zero results in a triangular path for the centroid of the top plate, which is spanned in a counterclockwise direction (Fig.~\ref{fig:fig3}d). Importantly, when we apply this pressure history to a physical sample, the actuator's tip follows the predicted trajectory and approaches the two targets very closely (see Fig.~\ref{fig:fig3}d), confirming the validity of our design strategy. As predicted by our model, the actuator starts straight, bends toward $T1$, bends toward $T2$ and finally returns to the straight configuration $O$ after reset.

For the second set of red targets $(T1,T2,T3)$, the larger height and radius of action of the targets lead to an optimal actuator comprising \mbox{$n_s = 4$} super-cells ({note that the convex shape of  $\Psi$ is due to a correlation between the optimal number of units and the average distance of the targets from the origin---see Supplementary Materials Fig.~\textcolor{blue}{S19})}. As shown in Fig.~\ref{fig:fig3}b, this 12-unit actuator comprises the classic Kresling module as well as bistable units with $\Delta=2$, $3$, and $4$ mm. This results in $8$ stable states, $14$ snapping transitions, and a more complex state diagram in which not all targets are reached consecutively by continuously decreasing pressure (Fig.~\ref{fig:fig3}e). More specifically, to move from $T1$ to $T2$, this actuator has to be reset by decreasing the pressure below $p_3^-$ before increasing above $p_4^+$ and then lowering it to $p_3^-$. As such, in this case the centroid of the top plate of the actuator passes through the straight configuration $O$ when moving from $T1$ to $T2$ and its trajectory  comprises two disconnected loops, $O-T1$ and $O-T2-T3$ (Fig.~\ref{fig:fig3}f). Note that we can add additional constraints to our greedy algorithm to make sure the targets fall within the same closed loop on the state diagram. This obviously leads to a different design and may increase the targets error (see Supplementary Materials Fig.~\textcolor{blue}{S17} for details). However, the ability to fully control the actuator's sequential trajectory makes the platform ideal for robotic applications.

\section*{MuA-Ori for robotic applications}

\begin{figure}[b!]
\centering
\includegraphics[width=1\linewidth]{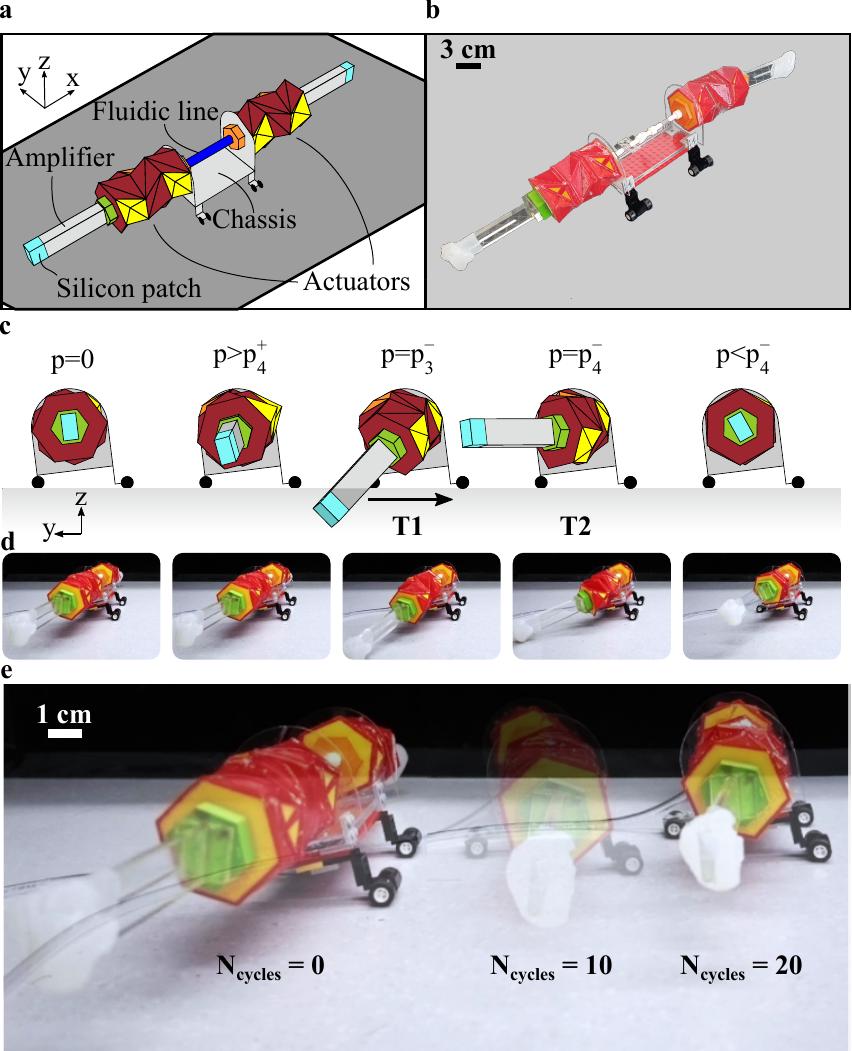}
\caption{\textbf{Land rowing robot}. \textbf{(a)} Schematics of the robot. \textbf{(b)} Robot prototype. \textbf{(c)-(d)} Trajectory followed by the robot's arms upon inflation and deflation, as predicted by the model (c) and observed in experiments (d). \textbf{(e)} Experimental snapshots of the robot in the initial configuration and after 10 and 20 cycles. }
\label{fig:fig4}
\end{figure}

To show the potential of the MuA-Ori for robotic applications we take inspiration from the closed trajectory of a rowing stroke, and use the closed triangular loop of the $3$-unit actuator in Fig.~\ref{fig:fig3}e to propel a land rowing robot (Figs.~\ref{fig:fig4}a). The robot comprises two symmetric arms (corresponding to the $3$-unit MuA-Ori in Fig.~\ref{fig:fig3}) connected through a single fluidic line and mounted on a wheeled chassis. Further, to harnesses the cyclic motion of the actuator and generate positive mechanical work with the ground, we connect two rigid rods
to the outer caps that serve as stroke amplifiers  and attach silicon patches at their ends to increase friction with the ground. When increasing the pressure above $p_4^+$ and then lowering it to $p_3^-$ the two actuators reach $T1$.  At this point the rigid rods touch the ground. Then, when we further decrease the pressure, the rigid rods move backward to T2 creating a forward thrust and eventually lift off from the ground, completing the stroke. Finally, lowering the pressure below $p_4^-$ resets the locomotion cycle (see Figs.~\ref{fig:fig4}c and d). As shown in the experimental snapshots in  Fig.~\ref{fig:fig4}e we can harness this particular trajectory instructed by the model to create locomotion: the robot advances of about $16$ cm in $20$ cycles.  Note that, differently from other robotic platforms with similar performance but requiring one or more actuators per leg with dedicated pressure sources \cite{multigait,Bootheaat1853}, our robots operate with a single pressure input, which largely simplifies its control.
\section*{Conclusion}

To summarize, in this work we have presented a platform to design fluidic origami robots that can switch between a range of different configurations  using only one pressurized input: the MuA-Ori. The key characteristics of our platform is an origami building block with a {degree-four vertex} panel, which can be geometrically programmed to snap at a certain input threshold, inducing  bending. This, together with the position of the {modified} panel in the origami module and their direction of rotation, constitute the parameters of a rich design space that we can efficiently scan with a custom greedy algorithm. Therefore, instructed by this model, we are able to automatically design and build optimal robotic arms that can reach a set of defined targets, and actuators capable of tracing spacial trajectories in a programmed sequence, enabling mechanical tasks and functions. While in this study we have used a simple geometric model to identify optimal designs, a fully mechanical model may be needed in order to extend the proposed approach to larger designs for which the effect of gravity is not negligible. {In addition, the current design space could be further expanded through investigating the effect of other geometrical parameters (e.g. $l$, $h$, and $\alpha$) on the resulting deformation of the modules, as well as expanding the range of the considered values of $\Delta$. While this could lead to more complex deformation modes and enhanced functionality, a drawback is a more complex state diagram. This means that a given actuator might have to go through a longer loading history to reach some prescribed targets, increasing the operational time-span. A potential solution to this is to measure the volume at which the module snaps inward and outward, assume constant flow rate, and derive the time associated to each snapping transition. This time span could then be included as variable in the optimization algorithm, in order to find a design that reaches the target in the shortest possible time.} To conclude, we envisage that the MuA-Ori concept hereby presented will serve as future strategy for the design of fluidic origami robots able to overcome the unimodal deformation constraint, despite the simple actuation.

\medskip

\section*{Materials and Methods}
\small{
{Details of the design, materials, and fabrication methods are summarized in Supplementary Materials, Section \textcolor{blue}{S1}. The experimental procedure of the inflation with water to measure the pressure-volume curve is described in Supplementary Materials, Section \textcolor{blue}{S2}, along with additional experimental data.experimental data. Details on the numerical mode are provided in Supplementary Materials, Section \textcolor{blue}{S3}. Finally, the optimization algorithms used in this study are described in detail in Supplementary Materials, Section \textcolor{blue}{S4}, along with additional data.
}}

\section*{Acknowledgments}
\small{We thank Eder Medina for his assistance in the development of the fabrication methodology. \textbf{Author contributions:} A.E.F., D.M., B.G. and K.B. proposed and developed the research idea. A.E.F designed and fabricated the samples. A.E.F., D.M., B.G. and L.K. performed the experiments. A.E.F. and D.M. designed the optimization. D.M. conducted the numerical calculations. A.E.F., D.M., and K.B. wrote the paper. K.B. supervised the research. \textbf{Funding:} Research was supported by the NSF grants DMR-2011754, DMR-1922321 and EFRI-1741685.
A.E.F. acknowledges that this project has received funding from the European Union’s Horizon 2020 research and innovation program under the Marie Skłodowska-Curie grant agreement No 798244. \textbf{Competing interests:} The authors declare no conflict of interest. \textbf{Data and materials availability:} All data needed to evaluate the conclusions of this study are present in the paper or the Supplementary Materials.
}

\section*{Supplementary materials}
\small{Section S1.\ Fabrication}\\
\small{Section S2.\ Testing}\\
\small{Section S3.\ Model}\\
\small{Section S4.\ Optimization}\\
\small{Figure S1.\ 3D-printed origami modules}\\
\small{Figure S2.\ Multi-unit actuator fabrication and assembly}\\
\small{Figure S3.\ Single pressure input origami robot}\\
\small{Figure S4.\ Experimental setup of the inflation with water}\\
\small{Figure S5.\ Experimentally measured response of the $\Delta=3$ mm module}\\
\small{Figure S6.\ Experimental pressure, displacement, and bending angle curves of our origami modules}\\
\small{Figure S7.\ Effects of $\Delta$}\\
\small{Figure S8.\ Modeling the stable states and snapping transitions of our origami modules}\\
\small{Figure S9.\ Actuator's parameters}\\
\small{Figure S10.\ State diagrams}\\
\small{Figure S11.\ A $2$-units actuator characterized by $[\Delta^1c^1f^1;\Delta^2c^2f^2] = [2\textrm{//} 3;4\textbackslash \textbackslash 3]$.}\\
\small{Figure S12.\ Design space for actuators with $n \in\{1,2,3,4\}$}\\
\small{Figure S13.\ Greedy algorithm}\\
\small{Figure S14.\ Comparison between optimization algorithms with integer constraints}\\
\small{Figure S15.\ Greedy algorithm results}\\
\small{Figure S16.\ A $6$-units actuator reaching three targets}\\
\small{Figure S17.\ The $12$-units actuator with additional constraint}\\
\small{Figure S18.\ Random targets error}\\
\small{Figure S19.\ Optimal number of units as a function of the target radius}\\
\small{Movie S1.\ Fabrication}\\
\small{Movie S2.\ 2-units actuators}\\
\small{Movie S3.\ 12-units actuator, triple target objective}\\
\small{Movie S4.\ Robotic application}\\

\section*{References}
\bibliography{references}

\clearpage
\newpage
\onecolumn
\section*{Supplementary Information}
\section*{S1. Fabrication}

The actuators tested in this study are constructed by connecting 3D-printed origami modules. This section gives details on the design and fabrication of the 3D-printed modules as well as their assembly to create multi-unit actuators and robots.

\subsection*{Design of the 3D-printed origami units}

\begin{figure}[h!]
\centering
\includegraphics[scale=0.99]{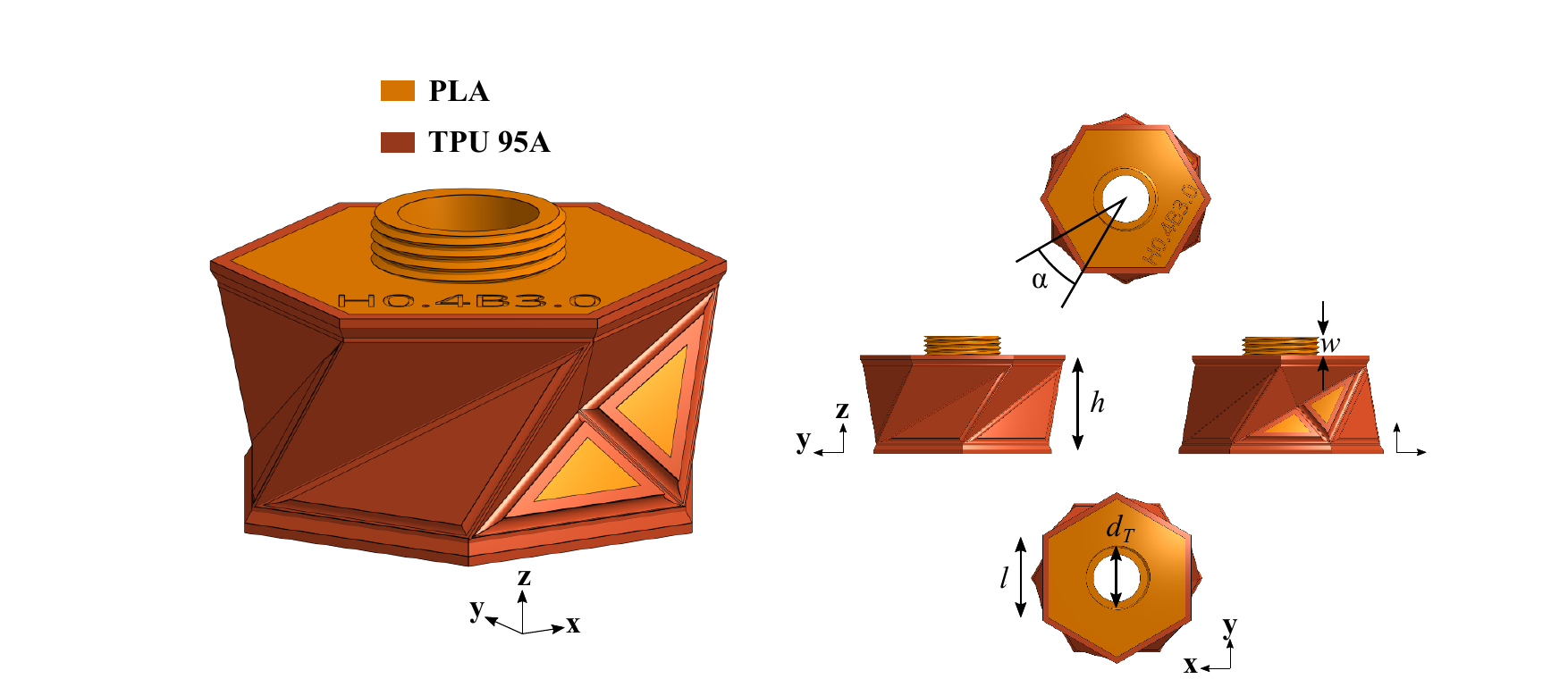}
\caption{\textbf{3D-printed origami modules.} Isometric and projected views of the origami module.  }\label{2_fabrication_fig0}
\end{figure}

Each origami module is 3D-printed using a commercially available multi-material printer (Ultimaker $3$). To account for geometric incompatibility during inflation, we print the $1$-mm thick triangular facets out of compliant thermoplastic polyurethane (Ultimaker TPU 95A with tensile modulus $E=26$ MPa). The thickness decreases to $0.4$ mm at the junction of the triangular facets (hinges), to allow more compliance. This value is the lowest possible thickness our printer is able to print with a $0.4$ mm print core.
Further, to enable rigid connection of different units and increase bistability, we print the end caps as well as the four triangular facets of the bistable cell out of stiff polyactic acid (Ultimaker PLA with tensile modulus $E=2.3$ GPa). As shown in Fig~\ref{2_fabrication_fig0}, the single module consists of two hexagonal caps with edges of length $l = 30$ mm, separated by a distance $h = 24$ mm, and rotated by an angle $\alpha = 30^\circ$ with respect to each other. To enable coupling of different units, we print a screw and a threaded hole on the top and bottom surfaces with length $w = 6$ mm an thread size $d_T = 24$ mm.

\subsection*{Assembly of an inflatable multi-unit actuator}

\begin{figure}[h!]
\centering
\includegraphics[scale=0.99]{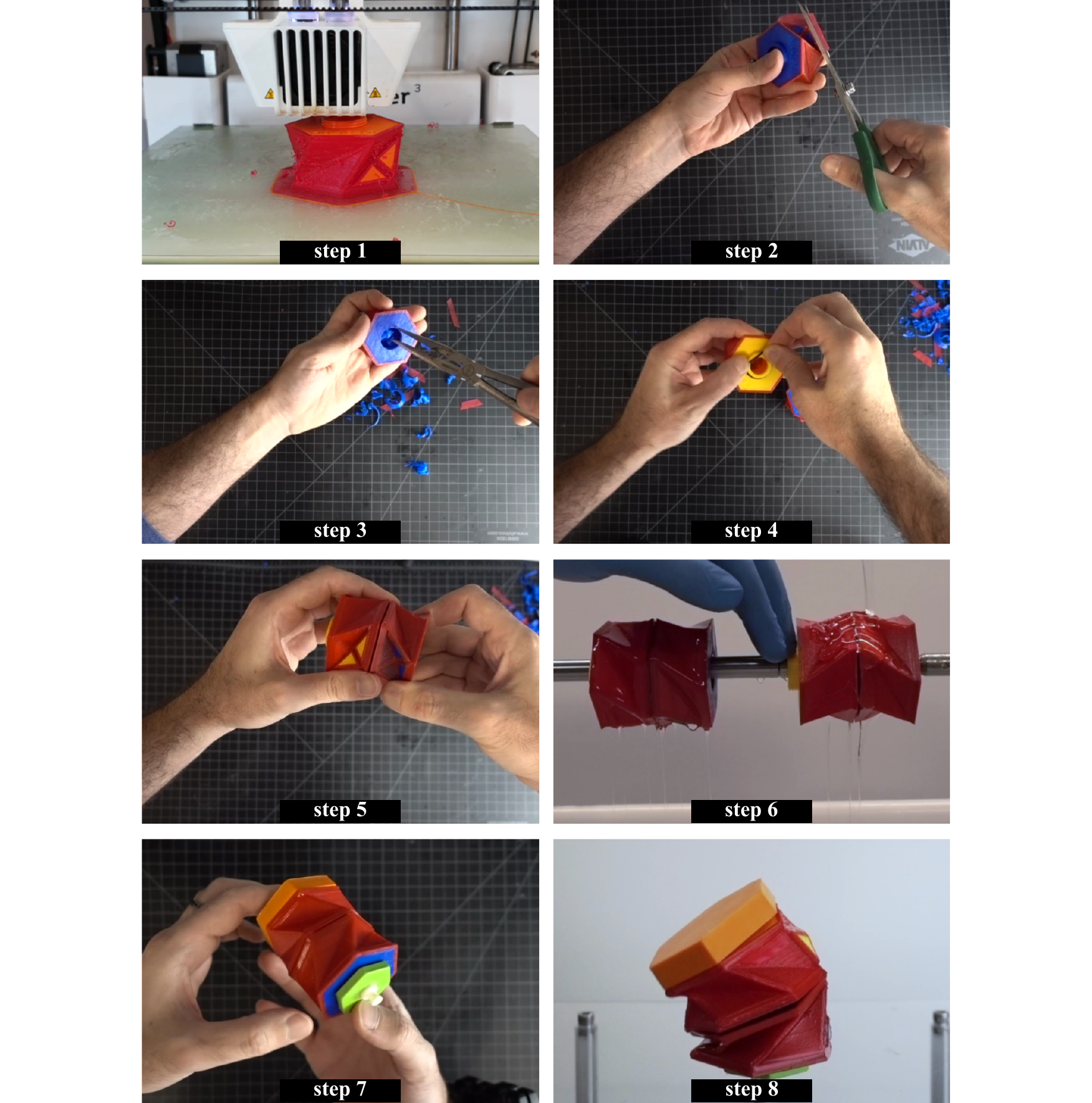}
\caption{\textbf{Multi-unit actuator fabrication and assembly.} Snapshots of the eight steps required to fabricate and assemble inflatable multi-unit origami actuators.}\label{2_fabrication_fig1}
\end{figure}

Below are the eight steps needed to fabricate and assemble an inflatable origami actuator sample made of multiple modules (see Fig. \ref{2_fabrication_fig1} and Movie \textcolor{blue}{S1}):

\begin{itemize}
\item {\textbf{Step 1:} we 3D-print (Ultimaker $3$) each origami unit out of polyactic acid (Ultimaker PLA) and thermoplastic polyurethane  (Ultimaker TPU 95A), using $0.4$ mm print cores with the \textit{fine} default setting.}
\item {\textbf{Step 2:} we cut the 3D-printed adhesion skirt with scissors.}
\item {\textbf{Step 3:} we remove the 3D-printed support material inside of the origami unit with pliers.}
\item {\textbf{Step 4:} we insert a toric joint on the connection screw to make the unit airtight (see \textbf{Steps 5-8}).}
\item {\textbf{Step 5:} we assemble multiple units together through the connection screws ensuring a tight assembly through the toric joints inserted in \textbf{Step 4}.}
\item {\textbf{Step 6:} we coat the sample with a $0.5$ mm layer of polydimethylsiloxane (PDMS) and let it cure for $24$ hours.}
\item {\textbf{Step 7:} we fit end caps, making sure to have a tight assembly through the toric joints inserted in \textbf{Step 4}, to create an airtight cavity. Note that one of the end caps has an inlet for fluidic actuation.}
\item {\textbf{Step 8:} we test the origami actuator by connecting it to a fluidic supply.}
\end{itemize}

\newpage
\subsection*{Design of the single pressure input origami robot}

\begin{figure}[h!]
\centering
\includegraphics[scale=0.99]{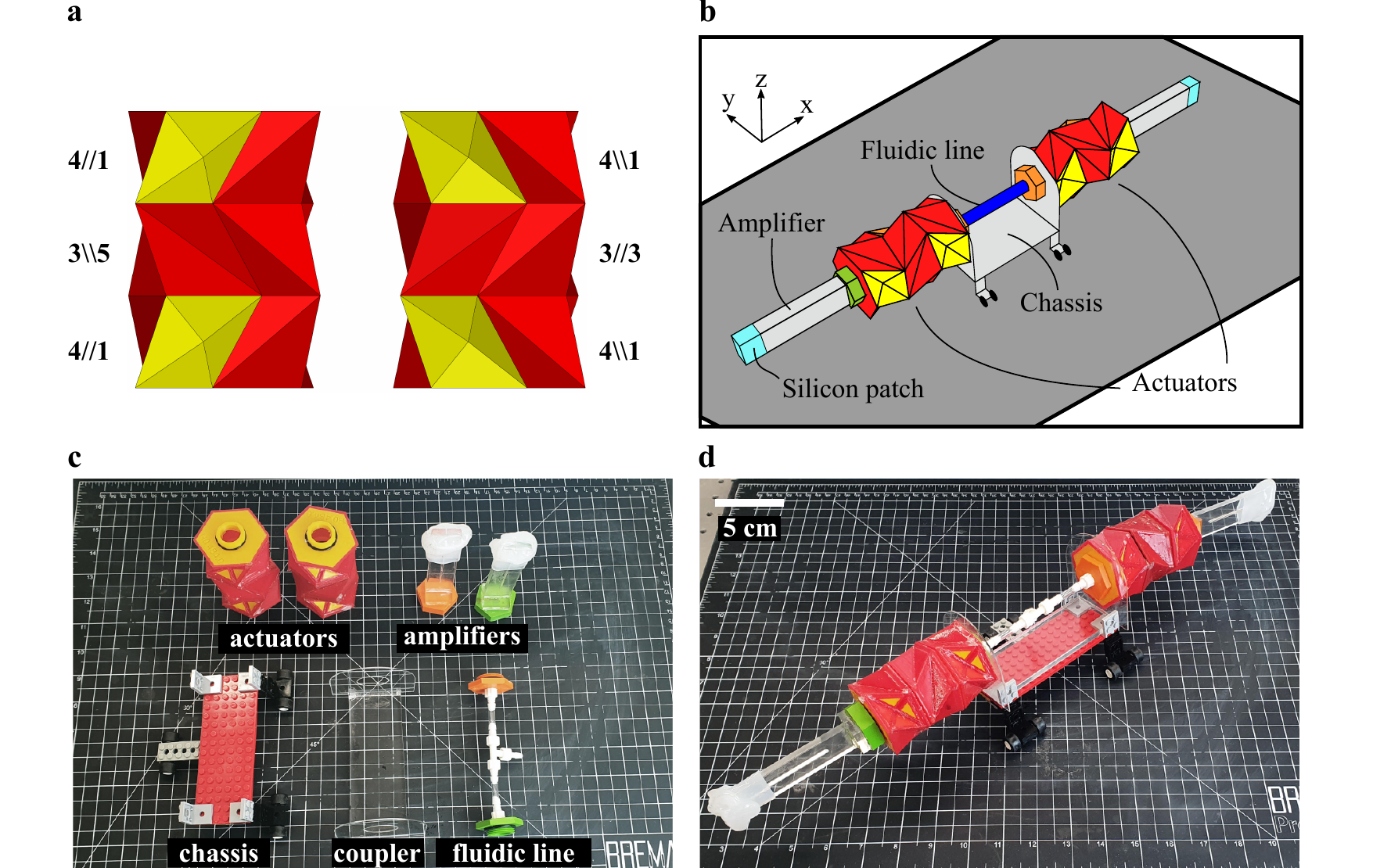}
\caption{\textbf{Single pressure input origami robot.} \textbf{(a)} The two identical actuators forming the arms of the robot. \textbf{(b)} Schematics of the origami robot. \textbf{(c)} Components of the origami robot. \textbf{(d)} Assembled origami robot.}\label{2_fabrication_fig3}
\end{figure}

The origami robot is constructed by mounting two multi-unit actuators (each one acting as a crawling arm) on a wheeled chassis made of LEGO bricks (see Figs.~\ref{2_fabrication_fig3}a and b). To create a single airtight cavity, we connect the two multi-unit actuators through a fluidic line (1/16" soft plastic tubing and luer lock connectors) and a coupler (1/16" acrylic plate fixed on the chassis). Finally, we glue $10$ cm-long amplifiers (1/4" acrylic plate) on the top caps of the actuators to accentuate the stroke during inflation and deflation. Note that we add a patch of silicon rubber on the amplifiers' tip (ecoflex 5 from Smooth-on) to increase the friction coefficient when in contact with the ground.

\clearpage
\newpage
\clearpage
\newpage
\section*{S2. Testing}

\begin{figure}[bh!]
\centering
\includegraphics[scale=0.75]{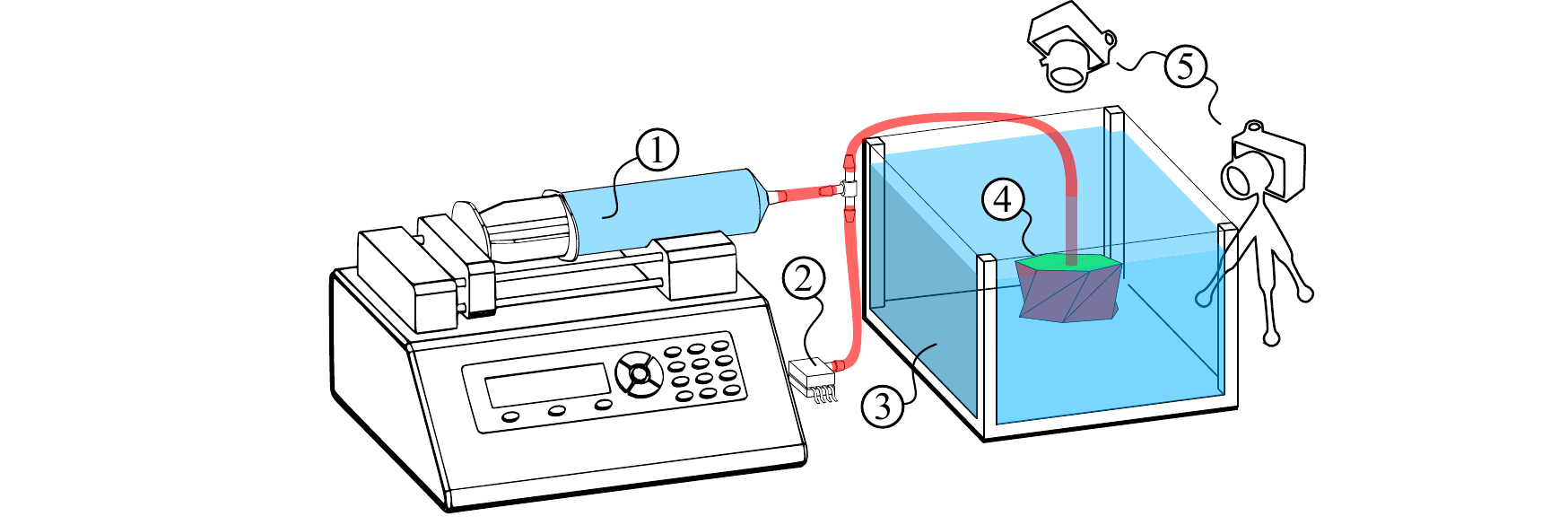}
\caption{\textbf{Experimental setup of the inflation with water.} Schematic of the test setup used to characterize the bending angle vs./ pressure and height vs./ pressure curves of the origami units with (1) syringe pump, (2) pressure sensor, (3) water tank, (4) origami unit, and (5) digital cameras.}\label{3_testing_fig1}
\end{figure}

To characterize the experimental response of the fabricated origami building blocks, we inflate them with water--- to eliminate the influence of air compressibility--- and measure their internal pressure while tracking their height and bending angle. As shown in Fig.~\ref{3_testing_fig1}, we use a syringe pump (Pump 33DS, Harvard Apparatus) to displace water into the origami unit at \mbox{$10$ mL/min}, measure the pressure using a pressure sensor (ASDXRRX015PDAA5 with a measurement range of $\pm 15$ psi by 	
Honeywell), and track the height, bending angle, and twist angle of the upper cap using two digital cameras (front and top view with two SONY RX100 V). Note that we submerge the entire unit in a water tank to eliminate the influence of gravity while eliminating air from all supply tubes and calibrate the pressure sensor to atmospheric pressure before each measurement cycle.

\begin{figure}[bh!]
\centering
\includegraphics[scale=1]{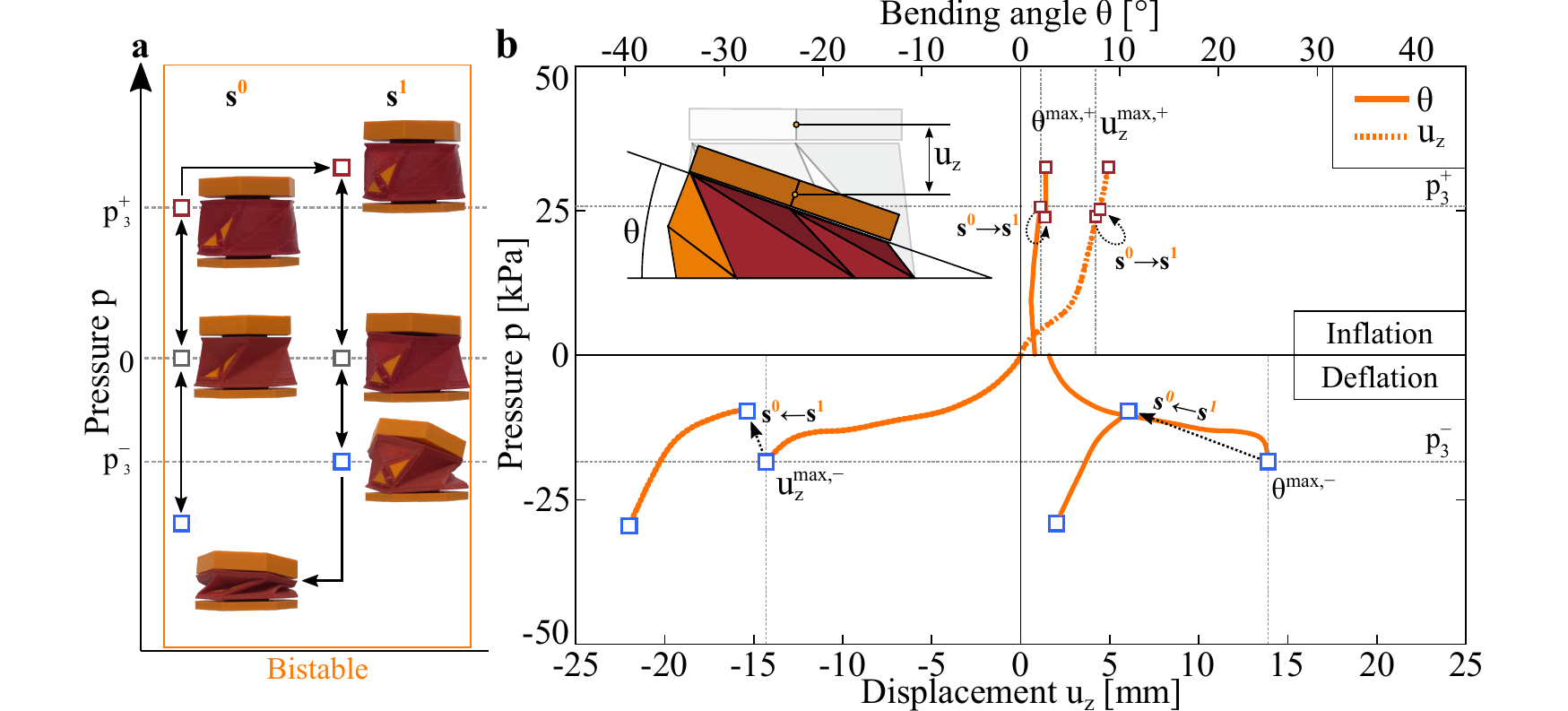}
\caption{\textbf{Experimentally measured response of the $\Delta=3$ mm module.} (a) State diagram of the pressurized origami module. (b) Experimentally measured pressure, $p$, as a function both end caps' displacement, $u_z$, and bending angle, $\theta$ for the origami module with $\Delta = 3$ mm. In the plot, we highlight  the pressure thresholds, $p_3^{+}$ and $p_3^{-}$, as well as the maximum displacement, $u_z^{max,\pm}$, and bending angle, $\theta_{max,\pm}$, at snap-through for both  inflation and deflation.}\label{3_testing_fig1c5}
\end{figure}

\noindent We report in Fig.~\ref{3_testing_fig1c5} the experimentally measured pressure, $p$, as a function of the end caps displacement, $u_z$, and the bending angle, $\theta$, for the module with $\Delta = 3$mm. Further, we highlight on the curves the pressure thresholds, $p_3^{+}$ and $p_3^{-}$, as well as the maximum displacement, $u_z^{max,\pm}$, and bending angle, $\theta_{max,\pm}$, at snap-through for both the inflation and deflation regime. 

\begin{figure}[H]
\centering
\includegraphics[scale=1]{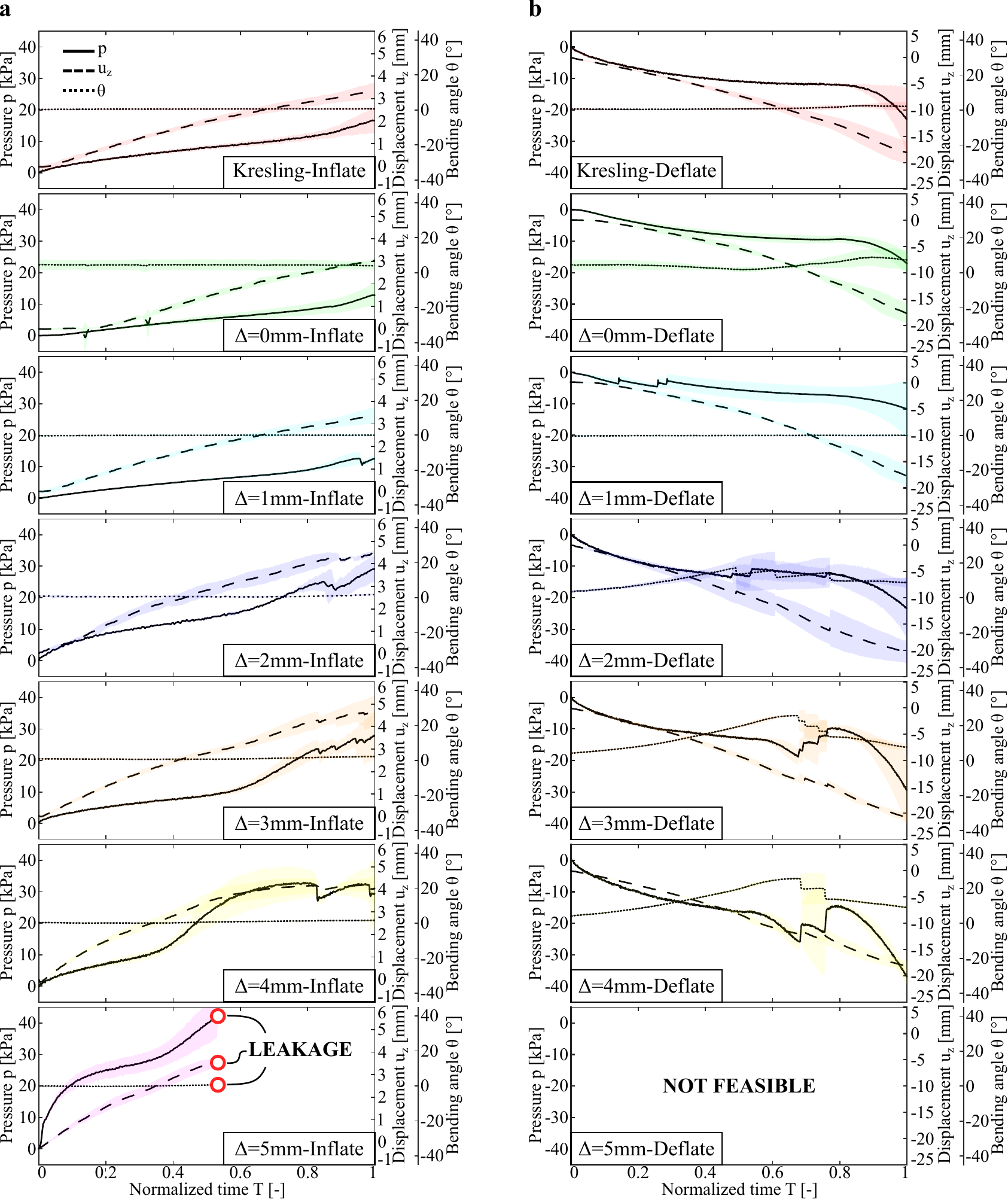}
\caption{\textbf{Experimental pressure, displacement, and bending angle curves of our origami modules.} . Experimentally measured pressure (left vertical axis), end caps displacement (right vertical axis), and bending angle (right vertical axis) of each design tested in this study as a function of normalized time, $T = t/t_{end}$ (where $t$ is real time and $t_{end}$ the duration of the test), for both (a) inflation and (b) deflation. To validate repeatability, we test for each design three specimens and report the mean (solid lines) and standard deviation (shaded region). We find that the classic Kresling module as well as the modified-panel module with $\Delta = 0$ mm do not show any snap-through instability. Modules with $\Delta = 1$ mm show a snapping transition during inflation, but bend marginally when later deflated. Modules with $\Delta = 2$, $3$, $4$ mm are bistable as they exhibit discontinuity in their $p-T$, $u_z-T$, and $\theta-T$ curves, and show substantial bending when deflated from the snapped configuration. Modules with $\Delta = 5$ mm break during the inflation before the panel snaps outward.}\label{3_testing_fig2}
\end{figure}

\noindent We summarize all experimental results in Fig.~\ref{3_testing_fig2} and plot the pressure, displacement, and bending angle of each design tested in this study as a function of normalized time, $T$, for both the inflation and deflation regime. To validate repeatability, we test for each design three specimens and report the mean (solid lines) and standard deviation (shaded region). We find that the classic Kresling module as well as the modified-panel module with $\Delta = 0$ mm do not show any snap-through instability. Modules with $\Delta = 1$ mm show a snapping transition during inflation, but bend marginally when later deflated. Modules with $\Delta = 2$, $3$, $4$ mm are bistable as they exhibit discontinuity in their $p-T$, $u_z-T$, and $\theta-T$ curves, and show substantial bending when deflated from the snapped configuration. Modules with $\Delta = 5$ mm break during the inflation before the panel snaps outward.

\begin{figure}[bh!]
\centering
\includegraphics[scale=1]{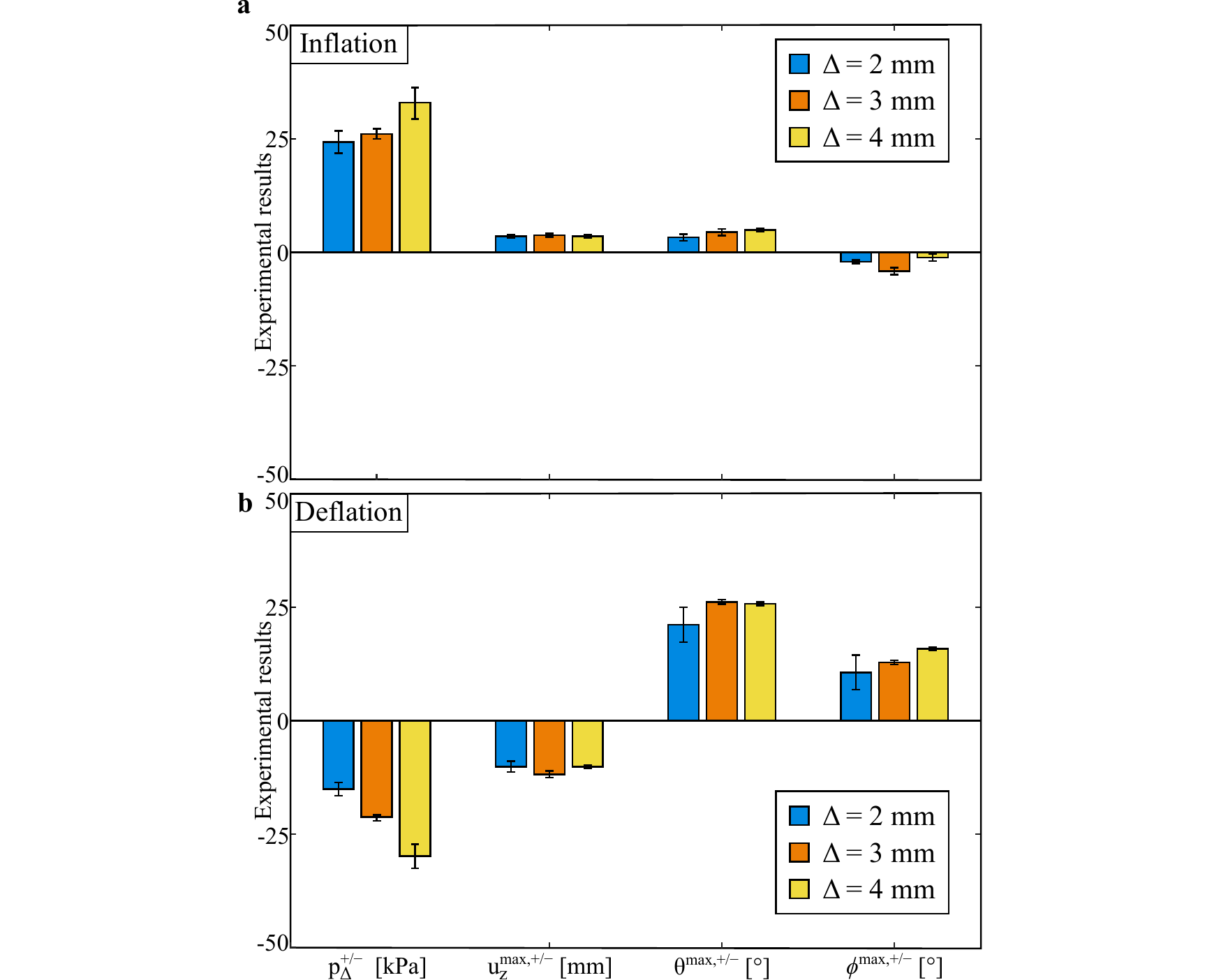}
\caption{\textbf{Effects of $\Delta$.} Bar chart of the experimentally measured pressure thresholds, $p_{\Delta}^{\pm}$, end caps displacement, $u_z^{max,\pm}$, bending angle, $\theta^{max,\pm}$, and twist angle, $\phi^{max,\pm}$, at snap-through  during both inflation \textbf{(a)} and deflation \textbf{(b)} The values for $p_{\Delta}^{\pm}$, $u_z^{max,\pm}$, and $\theta^{max,\pm}$ are extracted from the curves shown in Fig. \ref{3_testing_fig2}. The values for $\phi^{max,\pm}$ were obtained measuring the twist angle at each stable configuration and snapping transition during the experiments.}\label{3_testing_fig3}
\end{figure}

\newpage
\noindent From the experimental curves, we can extract the pressure thresholds at which the modified panel snaps outward during inflation, $p_\Delta^+$ and inward during deflation, $p_\Delta^-$. We summarize the mean and standard deviation of these pressure thresholds in Fig.~\ref{3_testing_fig3} for the bistable modules with $\Delta = 2$, $3$, and $4$ mm. We also report in Fig.~\ref{3_testing_fig3}, the displacement and the bending angle at snap-through, $u_z^{max,\pm}$ and $\theta_{max,\pm}$ for both the inflation and deflation regime.

%\noindent As an example, 

%Next, to summarize all experimental results, we plot in Fig.~\ref{3_testing_fig2}, the pressure, displacement, and bending angle of each design tested in this study as a function of normalized time, $T$, for both the inflation and deflation regime. To validate repeatability, we test for each design three specimens and report the mean (solid lines) and standard deviation (shaded region). We find that the classic Kresling module as well as the modified-panel module with $\Delta = 0$ mm do not show any snap-through instability. All other designs, i.e. modules with $\Delta = 2$, $3$, $4$ mm, are bistable as they exhibit discontinuity in their $p-T$, $u_z-T$, and $\theta-T$ curves. 

%From the experimental curves, we can extract the pressure thresholds at which the bistable panel snaps outward during inflation, $p_\Delta^+$ and inward during deflation, $p_\Delta^-$. We summarize the mean and standard deviation of these pressure thresholds in Fig.~\ref{3_testing_fig3} for the bistable modules with $\Delta = 2$, $3$, and $4$ mm. We also report in Fig.~\ref{3_testing_fig3}, the displacement and the bending angle at snap-through, $u_z^{max,\pm}$ and $\theta_{max,\pm}$ for both the inflation and deflation regime. 

\clearpage
\newpage
\section*{S3. Model}

\subsection*{Origami module}

\begin{figure}[bh!]
\centering
\includegraphics[scale=1]{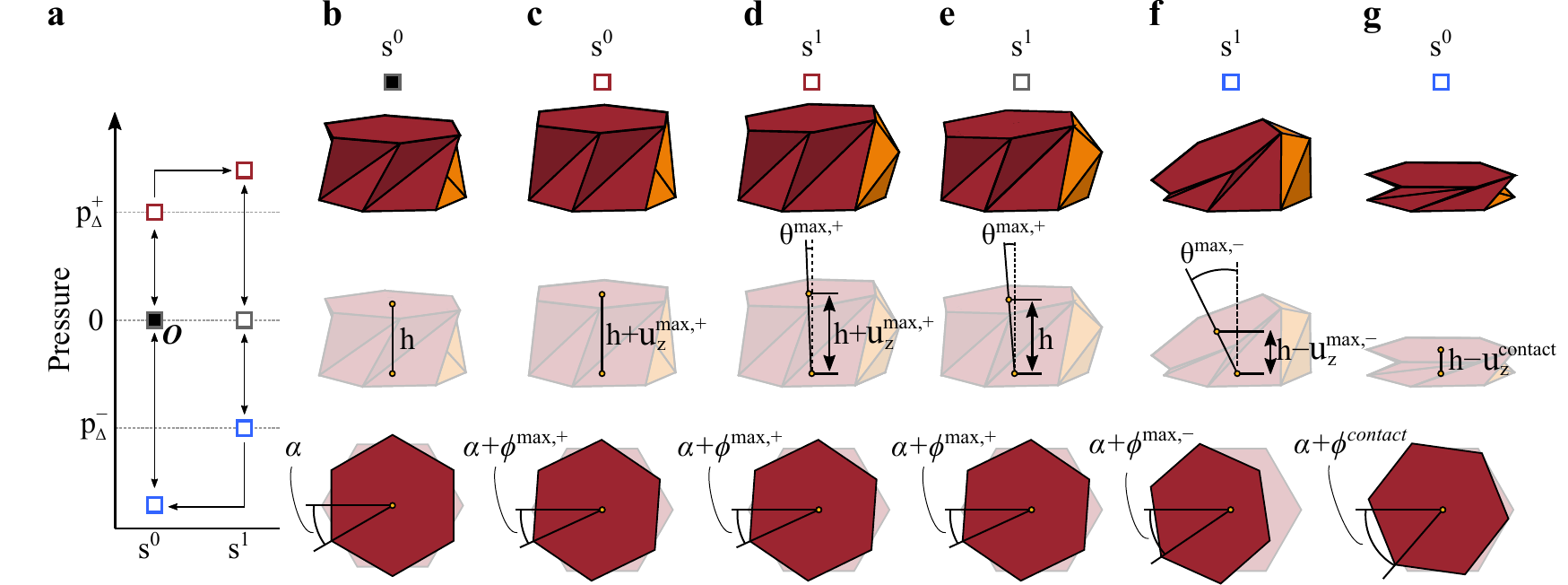}
\caption{\textbf{Modeling the stable states and snapping transitions of our origami modules.} \textbf{(a)} Typical state diagram for a origami module with a \textcolor{blue}{modified} panel of depth $\Delta$. The unit   can  transition between two stable states: state $s^0$ when the modified panel is folded inward, and state $s^1$ when the panel is popped outward and stays in that position even when the input pressure is removed \textbf{(b-g)} Reconstructed geometry of the origami module at each stable state and before and after each snapping transition.}\label{1_design_fig1}
\end{figure}

We start by using the geometric quantities that we track in our experiments (shown in Fig.~\ref{3_testing_fig3}) to reconstruct the configuration of the origami module just before and just after each snapping transition (see Fig.~\ref{1_design_fig1}).

\subsection*{Actuator comprising $n$ modules}

\begin{figure}[bh!]
\centering
\includegraphics[scale=1]{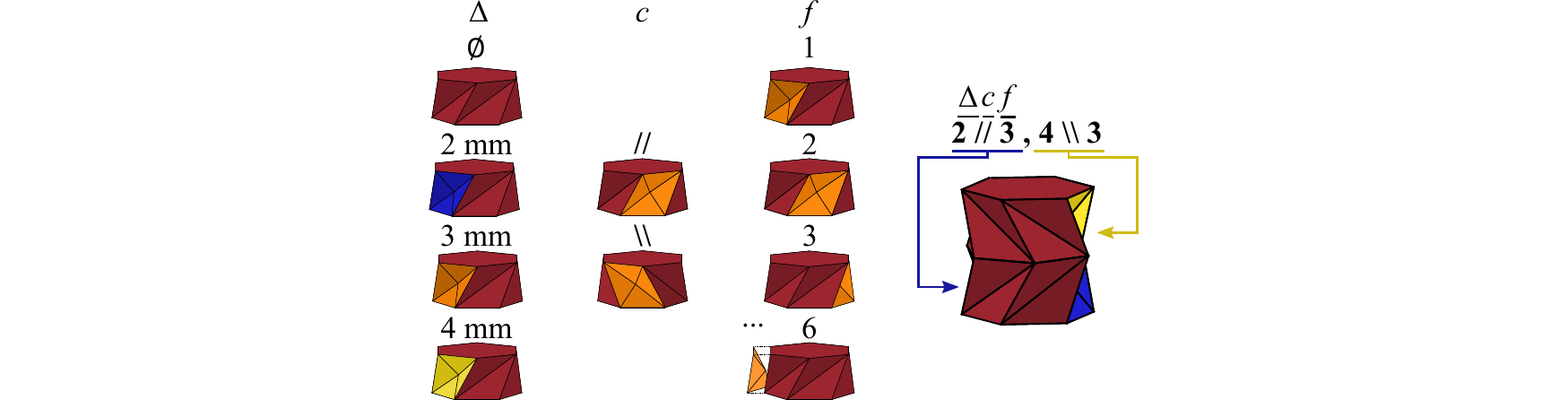}
\caption{\textbf{Actuator's parameters.} Our reconfigurable actuators are defined by the modified panel's depth, $\Delta^k$, the initial rotation of the upper cap, $c^k$, and the position of the modified panel, $f^k$, of each their $k$ unit.}\label{1_design_fig1c25}
\end{figure}

We can create actuator made of $n$ units by simply combining the different stable states and snapping transitions found for the single unit model described in Fig.~\ref{1_design_fig1}. Note that we impose that any $n$-units actuator forms a closed, inflatable cavity (i.e. they are all subjected to the same internal pressure). By combining $n$ modules, we can construct $\left(3\times2\times6+1\times2\right)^n$ different actuators since for each module $k$ we can select ($i$) either a regular Kresling module or a unit comprising a modified panel with depth $\Delta^k \in \{2,3,4\}$ mm; ($ii$) the upper cap to be rotated clockwise or anticlockwise with respect to the bottom one, $c^k\in \{\textrm{//},\textrm{\textbackslash \textbackslash}\}$,  and ($iii$) the side on which the modified panel is located, $f^k\in \{1,\ldots,6\}$ (see Fig.~\ref{1_design_fig1c25}).

\begin{figure}[bh!]
\centering
\includegraphics[scale=1]{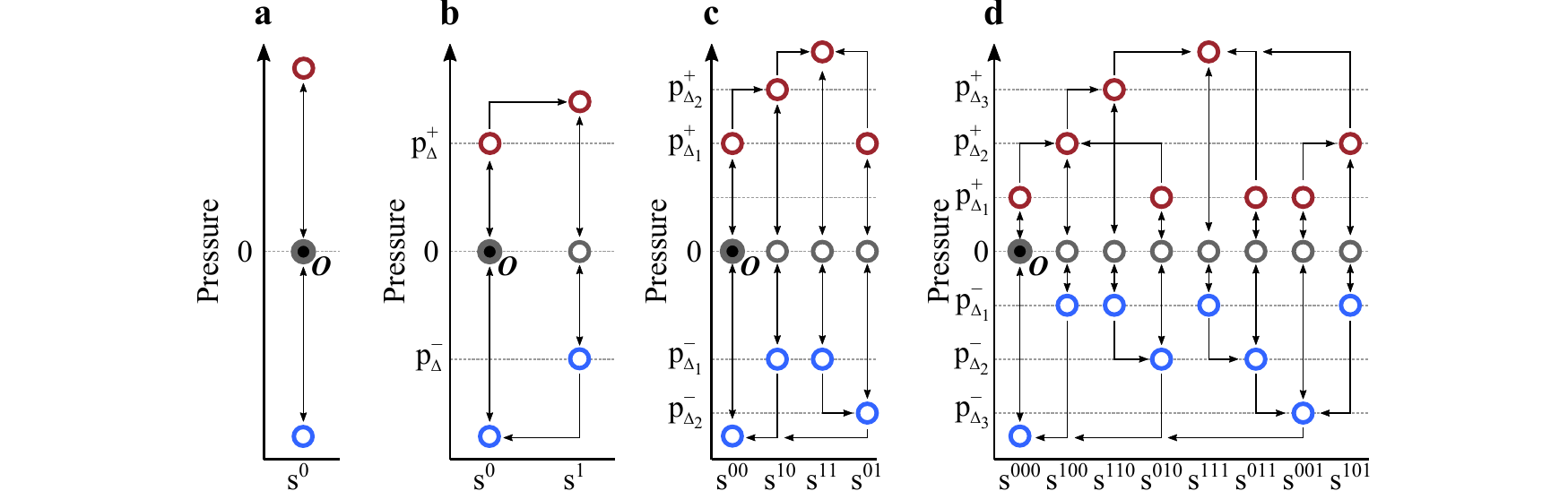}
\caption{\textbf{State diagrams.} \textbf{(a)} State diagram for any $n$-units actuator made only of Kresling modules. \textbf{(b-d)} State diagram for any $n$-units actuator made with $1$, $2$, and $3$ unique modified panel's depth, i.e. $n_\Delta = 1$, $2$, and $3$, respectively.}\label{1_design_fig1c5}
\end{figure}

\noindent For an actuator made of $n$ units, the number of stable states is equal to $2^{n_\Delta}$, where $n_\Delta$ is the number of unique modified panel depths $\Delta$. Note that we assume all units with the same $\Delta$ snap synchronously at the pressure thresholds, $p_\Delta^+$ and $p_\Delta^-$. Since in this study, we consider only the discrete set $\Delta \in \{2,3,4\}$ mm, all our actuators have either $n_\Delta = 0$, $1$, $2$, or $3$. For each different $n_\Delta$, we report the corresponding state diagram in Fig.~\ref{1_design_fig1c5}.\newpage

\begin{figure}[bh!]
\centering
\includegraphics[scale=0.9]{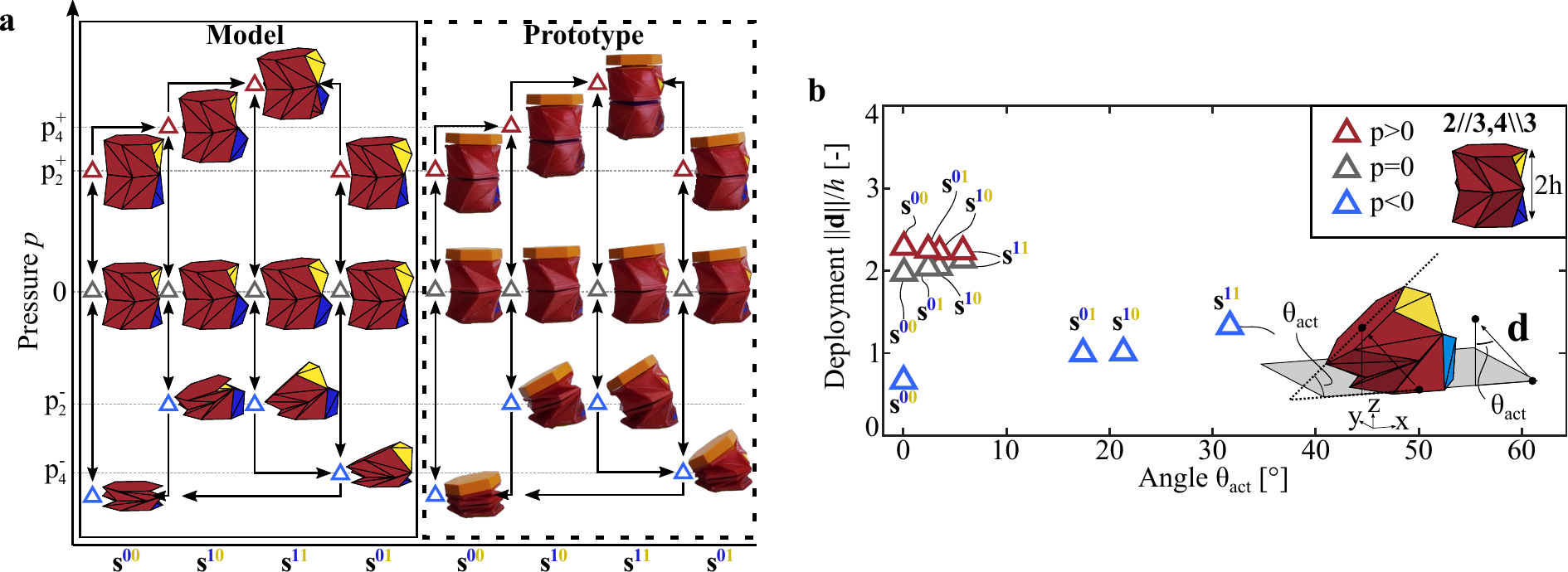}
\caption{\textbf{A $2$-units actuator characterized by $[\Delta^1c^1f^1;\Delta^2c^2f^2] = [2\textrm{//} 3;4\textbackslash \textbackslash 3]$.} \textbf{(a)} State diagram for a $2$-units actuator characterized by $[\Delta^1c^1f^1;\Delta^2c^2f^2] = [2\textrm{//}3;4\textbackslash \textbackslash 3]$: geometrical model prediction on the left, experimental snapshots on the right. \textbf{(b)} Normalized deployment, $||\mathbf{d}||/h$, as a function of the bending angle $\theta_{act}$, for every stable state and snapping transition.}\label{1_design_fig2}
\end{figure}

\begin{figure}[bh!]
\centering
\includegraphics[scale=1]{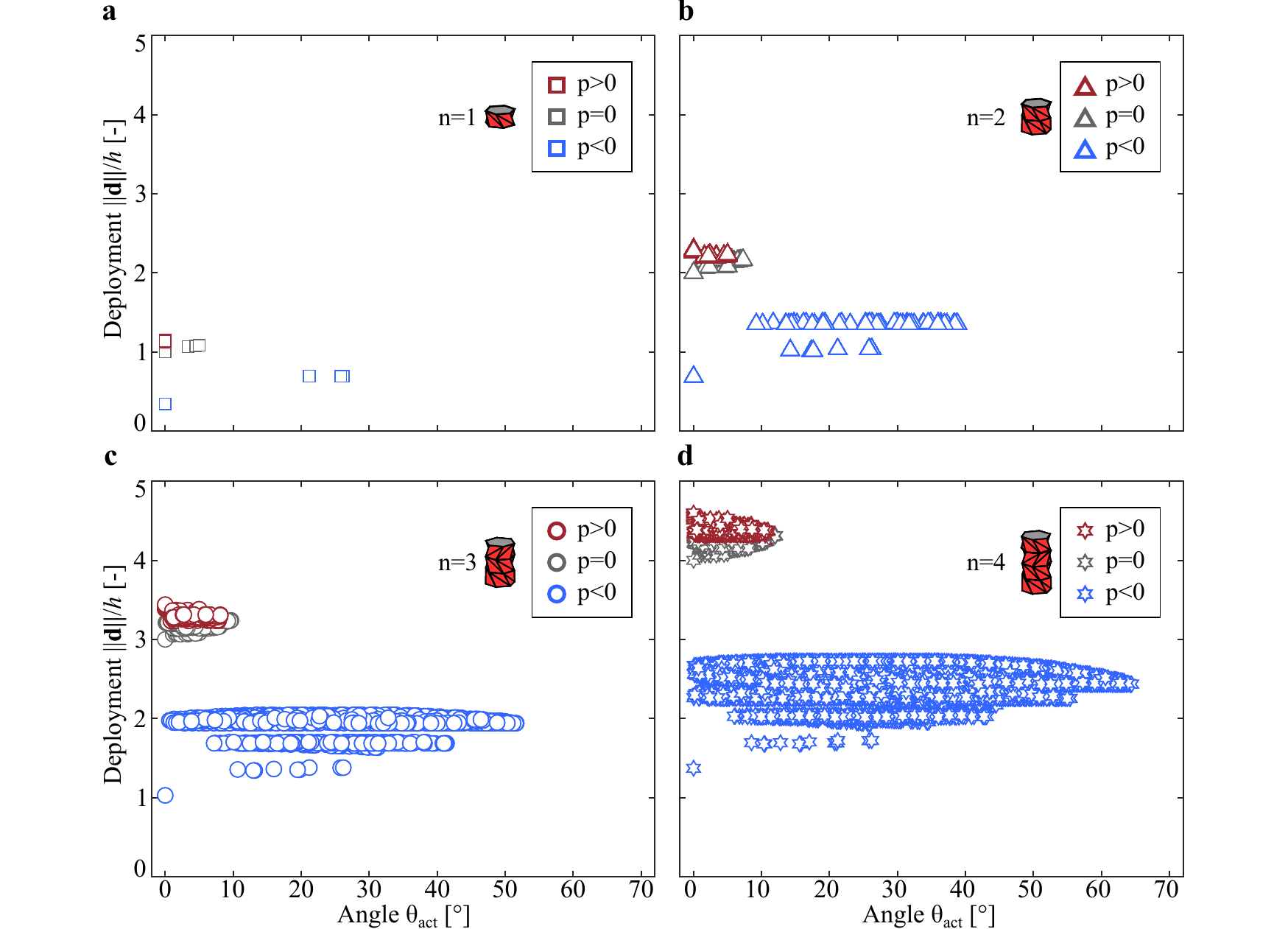}
\caption{\textbf{Design space for actuators with $n \in\{1,2,3,4\}$.} Normalized deployment, $||\mathbf{d}||/h$, and bending angle, $\theta_{act}$, for all possible $1$-unit \textbf{(a)}, $2$-unit \textbf{(b)}, $3$-unit \textbf{(c)}, and $4$-unit \textbf{(d)} constructed by combining our library of modules.}\label{1_design_fig3}
\end{figure}

\noindent In Fig.~\ref{1_design_fig2}a we show  the  state diagram of an actuators comprising   two modules  characterized by $[\Delta^1c^1f^1;\Delta^2c^2f^2] = [2\textrm{//} 3;4\textbackslash \textbackslash 3]$ and compare the outputs from the geometric model to the experiments done on a physical prototype. Further,
for each of the stable states and snapping transitions in the state diagram, we  record the vector connecting the caps' centroids, $\mathbf{d}$.   In Fig.~\ref{1_design_fig2}b, we report the normalized deployment, $||\mathbf{d}||/h$, and angle, $\theta_{act}$ for each stable state and snapping transition of the actuator characterized by $[\Delta^1c^1f^1;\Delta^2c^2f^2] = [2\textrm{//} 3;4\textbackslash \textbackslash 3]$. Further, in Fig.~\ref{1_design_fig3} we report  $||\mathbf{d}||/h$ and $\theta_{act}$  for each stable state and snapping transition of every actuator design with $n \in {1,2,3,4}$.

\newpage

\clearpage
\newpage
\section*{S4. Optimization}
To identify actuators capable of achieving target deformation modes, we solve the following discrete optimization problem 
\begin{equation}\label{minprob}
\begin{aligned}
\min_{\Delta^k,c^k,f^k} \quad & \Psi\left(\Delta^k,c^k,f^k;n\right)\\
\textrm{s.t.} \quad &\Delta^k\in \{2,3,4\}\textrm{mm}\\
 \quad &c^k\in \{\textrm{//},\textrm{\textbackslash \textbackslash}\} \\
  \quad&f^k\in\{1,2,3,4,5,6\}\\
  \quad&k \in \{1,2,\ldots,n\}\\
 \quad &n\in \mathcal{Z}^+,
\end{aligned}
\end{equation}
\noindent  where $\Psi$ is the cost function,  $n$ is the number of units making the actuator and $\Delta^k$, $c^k$, and $f^k$ are are the modified panel depth, the orientation of the upper cap with respect to the bottom one, and the side on which the modified panel is located for the $k$-unit in the array. Note that all design variables (i.e.  $\Delta^k$, $c^k$, and $f^k$) are constrained to be integer value and, for the sake of simplicity, we solve the optimization problem multiple times for fixed number of units $n \in [1,15]$.\\
\\
\noindent In the main text we use the optimization algorithm to identify actuators  whose tip can approach a desired set of target points and therefore define the cost function as
\begin{equation}\label{Psi}
    \Psi = \frac{1}{n_{targets}\cdot h}\sum_{m=1}^{n_{targets}}\min ||\mathbf{d} - \mathbf{T}_m||,
\end{equation}
where $n_{targets}$ is the number of targets, and $\mathbf{T}_m$ is the vector connecting the $m$-th target with the origin. Here, we also consider two additional optimization problems and use the algorithm to identify actuators that 
\begin{itemize}
\item maximize the angle $\theta_{act}$  between the vector connecting the cap's centroids, $\mathbf{d}$, and the $z$-axis. For this case we define the cost function as
\begin{equation}
    \Psi\left(\Delta^k,c^k,f^k;n\right) = -\theta_{act}.
\end{equation}
\item maximize the deployment height $||\mathbf{d}||/h$. For this case we define the cost function as
\begin{equation}
    \Psi\left(\Delta^k,c^k,f^k;n\right) = -||\mathbf{d}||/h,
\end{equation}
\end{itemize}
\subsection*{Optimization algorithms}
There are many algorithms able to solve an optimization problem with integer constraints such as the one presented in Eq.~[\ref{minprob}]. In this study, we used three classic algorithms: ($i$) the genetic algorithm with integer constraints; ($ii$) the integer optimization via a surrogate model;  and ($iii$) the greedy algorithm based on best-first search. Note that given the high-dimensionality and complexity of this optimization problem, there is no guarantee that these algorithms will lead to a unique global minimum. 
\subsubsection*{Genetic algorithm with integer constraints}
We started by using the genetic algorithm with integer constraints, which attempts to minimize a penalty function that depends on the fitness (value of the cost function $\Psi$) and feasibility (design variables are integer) of an individual. For this study, we used the Matlab implementation of the algorithm (Matlab function \texttt{ga}) and imposed the constraint that all design variables, i.e. $\Delta^k$, $c^k$, and $f^k$ must have integer values with upper and lower bounds reported in Eq.~[\ref{minprob}].  We ran the function \texttt{ga} multiple times, each time considering a fixed value of $n\in[1, 15]$, using a population size of $200$, a max stall generations (i.e. the consecutive number of generations with no change to the cost function value) of $500$ and a maximum number of generations of $1000$.  
\subsubsection*{Integer optimization via a surrogate model}
Next, we used the  surrogate model optimization, which is a derivative-free method that replaces the complex and non-smooth objective function by a surrogate (i.e. an approximation of that function), which is created by sampling the objective function. For this study we used the Matlab implementation of the algorithm (Matlab function \texttt{surrogateopt}) and imposed the constraint that all design variables, i.e. $\Delta^k$, $c^k$, and $f^k$, must have integer values with upper and lower bounds reported in  Eq.~[\ref{minprob}]. We ran the algorithm for fixed values of $n$ with a maximum number of function evaluations of $20,000$. 
\subsubsection*{Greedy algorithm based on best-first search}
\begin{figure}[H]
\centering
\includegraphics[scale=1]{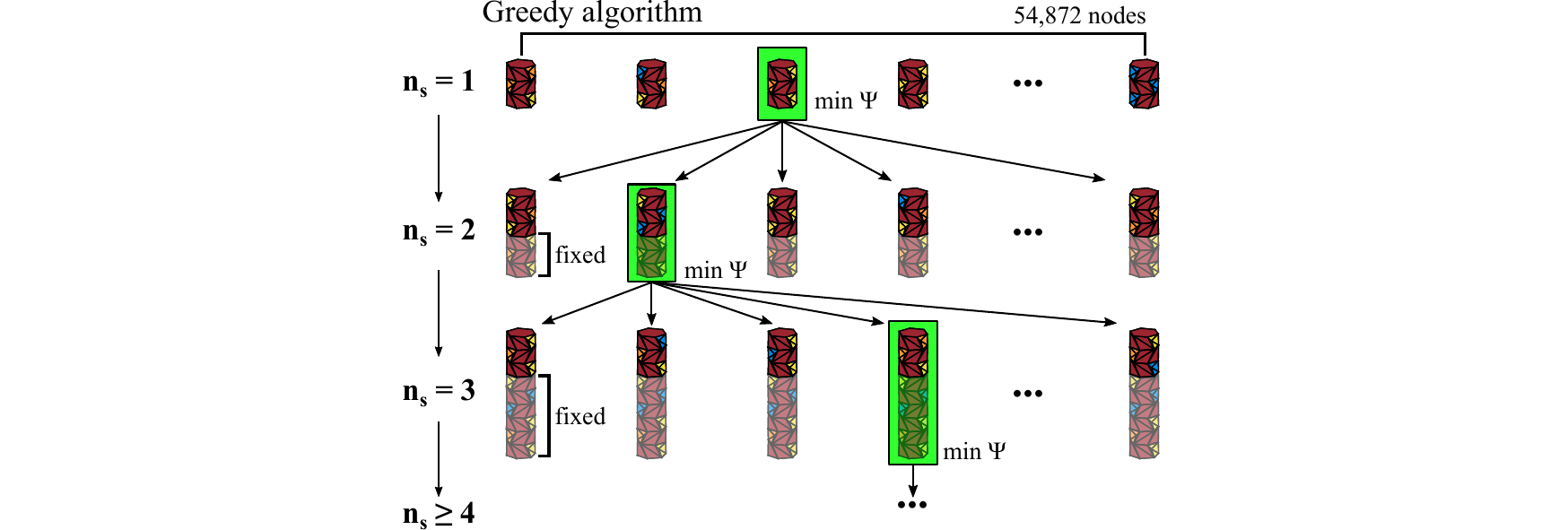}
\caption{\textbf{Greedy algorithm.} 
Schematic of the greedy algorithm with $n_u = 3$. At each iteration, the algorithm selects the actuator super-cell design that minimizes $\Psi$.}\label{fig_greedy_algo}
\end{figure}

For the greedy algorithm, we developed and in-house Matlab code based on the best-first search method that creates an actuator with $n$ units out of $n_s$ super-cells, each super-cell made of $n_u$ modules (so that $n = n_u \cdot n_s$).  At the first iteration, the algorithm selects the super-cell design that minimizes $\Psi$ and stores it in memory. Then, in the second iteration, we identify a second super-cell that, when connected to the first one, minimizes $\Psi$. The first two super-cells are then stored in memory and the algorithm advances to the next iteration (see Fig.~\ref{fig_greedy_algo} and Algorithm 1 below). 
\begin{algorithm}[H]\label{algo}
\caption{Greedy algorithm based on best-first search}
\label{algo1}
\textbf{Set} $n_{max}$;\\
\textbf{Set} $n_u$;\\
\textbf{Set} $n_s=0$;\\
\textbf{While} $n_s\cdot n_u \le n_{max}$\\
${}$\hspace{1em} $n_s = n_s + 1$;\\
${}$\hspace{1em} \textbf{if} $n_s =1$ \textbf{then}\\
${}$\hspace{2em} $\cdot$ Calculate $\Psi$ for each actuator design with ($\Delta^k,c^k,f^k$), $k=1:n_s\cdot n_u$;\\
${}$\hspace{2em} $\cdot$ Find the actuator that minimizes $\Psi$ and set its design variables to ($\Delta^{k^*},c^{k^*},f^{k^*}$)\\
${}$\hspace{1em} \textbf{else}\\
${}$\hspace{2em} $\cdot$ Calculate $\Psi$ for each actuator design with  ($\Delta^k,c^k,f^k$), where the set of variables from $k = 1:(n_s-1)\cdot n_u$ are coming\\
${}$\hspace{2em} from the previous iteration of ($\Delta^{k^*},c^{k^*},f^{k^*}$) and $k=(n_s-1)\cdot n_u + 1:n_s\cdot n_u$ are free.\\
${}$\hspace{2em} $\cdot$ Find the actuator that minimizes $\Psi$ and set its design variables to ($\Delta^{k^*},c^{k^*},f^{k^*}$)\\
%$\cdot$ Select the actuator design that minimizes $\Psi$. \KB{not sure we need this}
\end{algorithm}
\subsection*{Results} In the following we first compare the performance of the three algorithms and then present additional results obtained using the greedy algorithm.
\subsubsection*{Comparison between the three algorithms} 
\begin{figure}[H]
\centering
\includegraphics[scale=1]{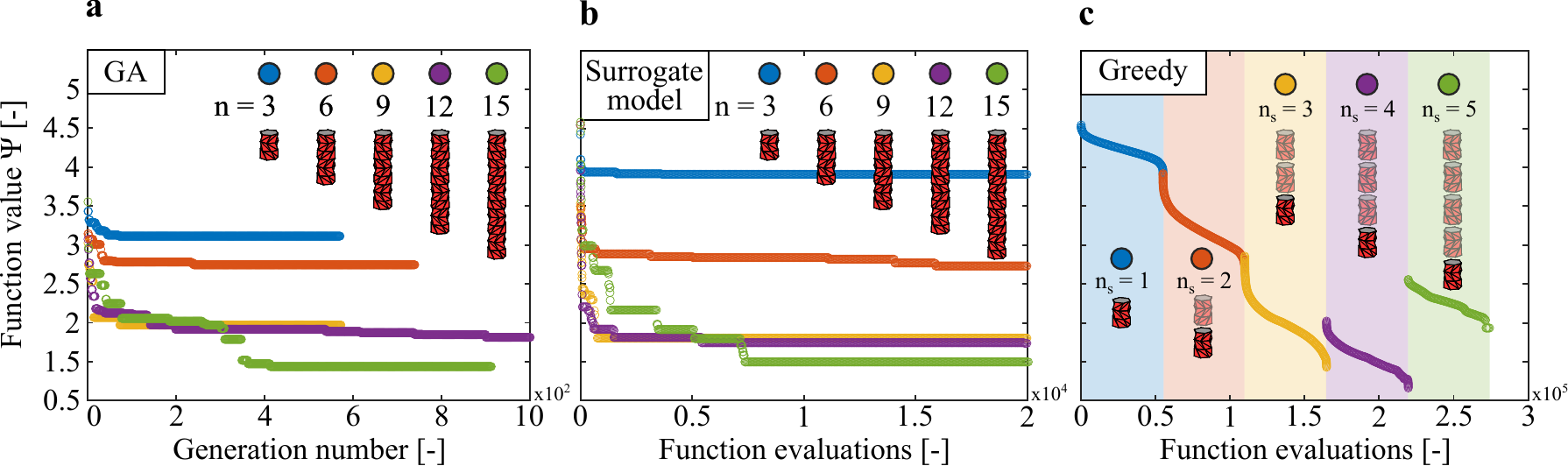}
\caption{\textbf{Comparison between optimization algorithms with integer constraints.} 
Comparison of the \textbf{(a)} generic algorithm, \textbf{(b)} surrogate model, and \textbf{(c)} greedy algorithm based on best-first search to solve the integer optimization problem of minimizing the targets error $\Psi$.}\label{fig_r_opt}
\end{figure}

To test and compare the three algorithms, we considered the set of three targets ($T_1$, $T_2$, $T_3$) shown in Fig.~\textcolor{blue}{3} of the main article and minimized the cost function given in Eq.~[\ref{Psi}]. In Fig.~\ref{fig_r_opt}, we report the cost function value with respect to the number of generations/function evaluations as obtained using the three algorithms.  We find that for all considered values of $n$ both the genetic algorithm with integer constraints and the surrogate model stall quickly, with a minimum value of the cost function of $1.04$ and $1.12$ reached for $n=15$, respectively. Further, we find that the greedy algorithm with $n_u=3$ outperforms the genetic algorithm and the surrogate model optimization as it identifies an actuator design that leads to $\Psi=0.729$ for $n =12$. Note that, for $n_u = 3$, the greedy algorithm requires about $2.75\times 10^5$ evaluations of $\Psi$ to identify the optimal design, whereas the surrogate model takes about $1\times 10^5$ evaluations of $\Psi$ (the genetic algorithms requires about $8\times10^5$ evaluations of $\Psi$ with a population size of $200$). However, the greedy algorithm does not require any other operation apart from a simple computation of $\Psi$ during each iteration. Differently, the surrogate algorithm has to update the underlying model. The simplicity of the greedy algorithm leads to a CPU time of $850$ s (parallelized on $24$ cores) to solve the algorithm compared to $2,500$ s and $4,000$ s for the genetic algorithm and the surrogate model, respectively. We therefore use the greedy algorithm to identify optimal configurations for our actuators.
\subsubsection*{Additional results generated by the greedy algorithm}
\begin{figure}[H]
\centering
\includegraphics[scale=1]{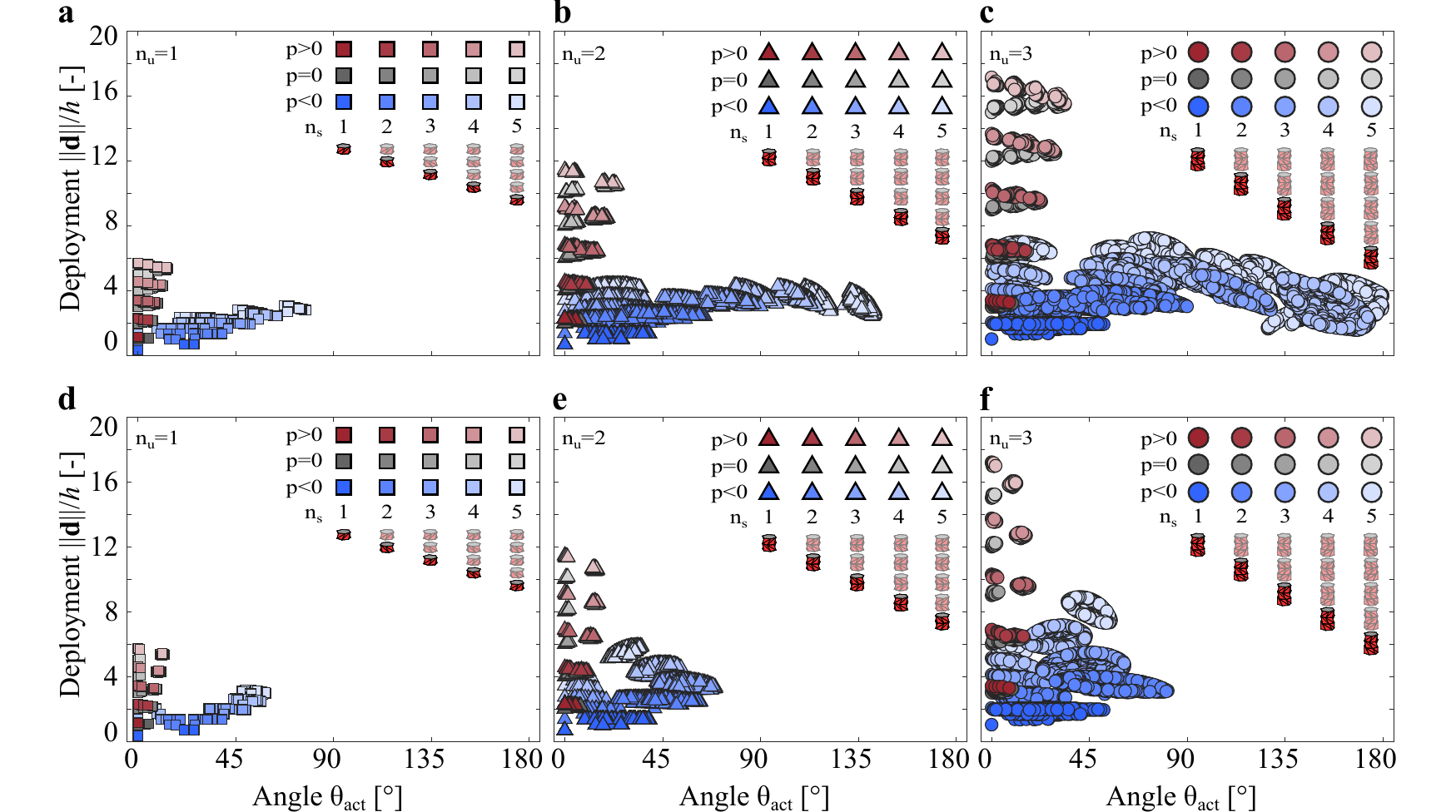}
\caption{\textbf{Greedy algorithm results}. Normalized deployment, $||\mathbf{d}||/h$, and bending angle, $\theta_{act}$, for actuators with $n_s \in\{1,2,3,4,5\}$ and $n_u\in\{1,2,3\}$. \textbf{(a-c)} The greedy algorithm is programmed to maximize $\theta_{act}$ at each increment of $n_s$. \textbf{(d-f)} The greedy algorithm is programmed to maximize $||\mathbf{d}||/h$ at each increment of $n_s$. }\label{1_design_fig4}
\end{figure}

\begin{figure}[H]
\centering
\includegraphics[scale=1]{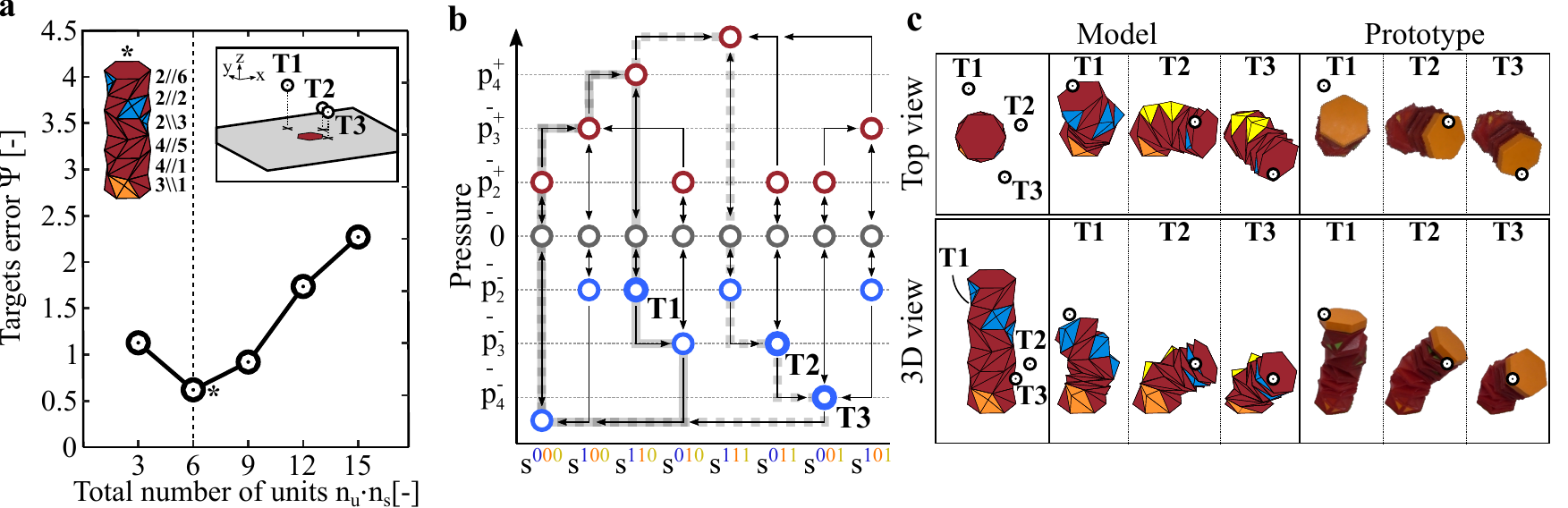}
\caption{\textbf{A $6$-units actuator reaching three targets.} For a specific set of three targets, we use the greedy algorithm to find the actuator design that minimizes $\Psi$, i.e~the error between the targets and the top cap's centroid. Note that we fix $n_u =3$ and consider $n_s\in\{1,2,3,4,5\}$. \textbf{(a)} Targets error $\Psi$ as a function of total number of units: the optimal actuator produced by the greedy algorithm for the three targets is reported as (*), along with the respective parameters for each module. The considered set of targets is shown in the inset. \textbf{(b)} State diagram for the $6$-units actuator (*) with targets $T1$, $T2$, and $T3$ highlighted. \textbf{(c)} Top and 3D view of the model and the experimental prototype for the $6$-units actuator reaching the targets. }\label{1_design_fig5}
\end{figure}

\begin{figure}[H]
\centering
\includegraphics[scale=1]{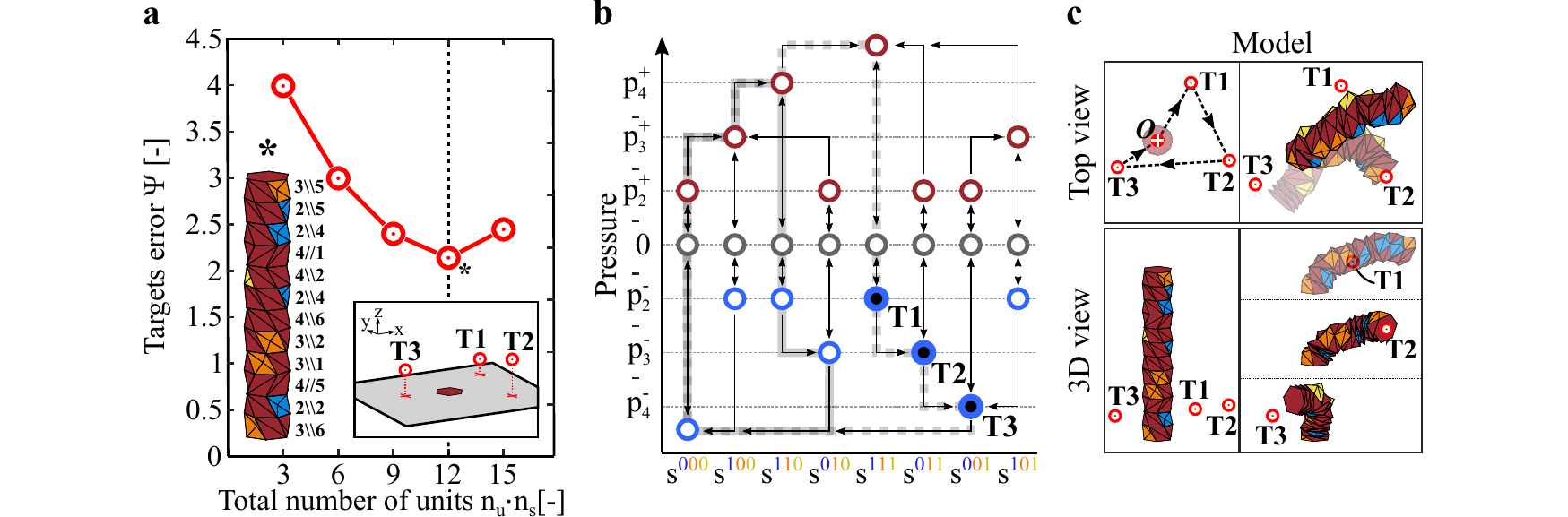}
\caption{\textbf{The $12$-units actuator with additional constraint.} We focus on the same set of three targets considered in Fig.~\textcolor{blue}{3} of the manuscript and further impose that each targets much be reached successively by decreasing pressure. \textbf{(a)} Targets error $\Psi$ as a function of total number of units: the optimal actuator produced by the greedy algorithm for the three targets is reported as (*), along with the respective parameters for each module. \textbf{(b)} State diagram for the optimal $12$-units actuator (*) with targets $T1$, $T2$, and $T3$ highlighted. \textbf{(c)} Top and 3D view of the model for the $12$-units actuator reaching the targets.}\label{1_design_fig7}
\end{figure}

\begin{figure}[H]
\centering
\includegraphics[scale=1]{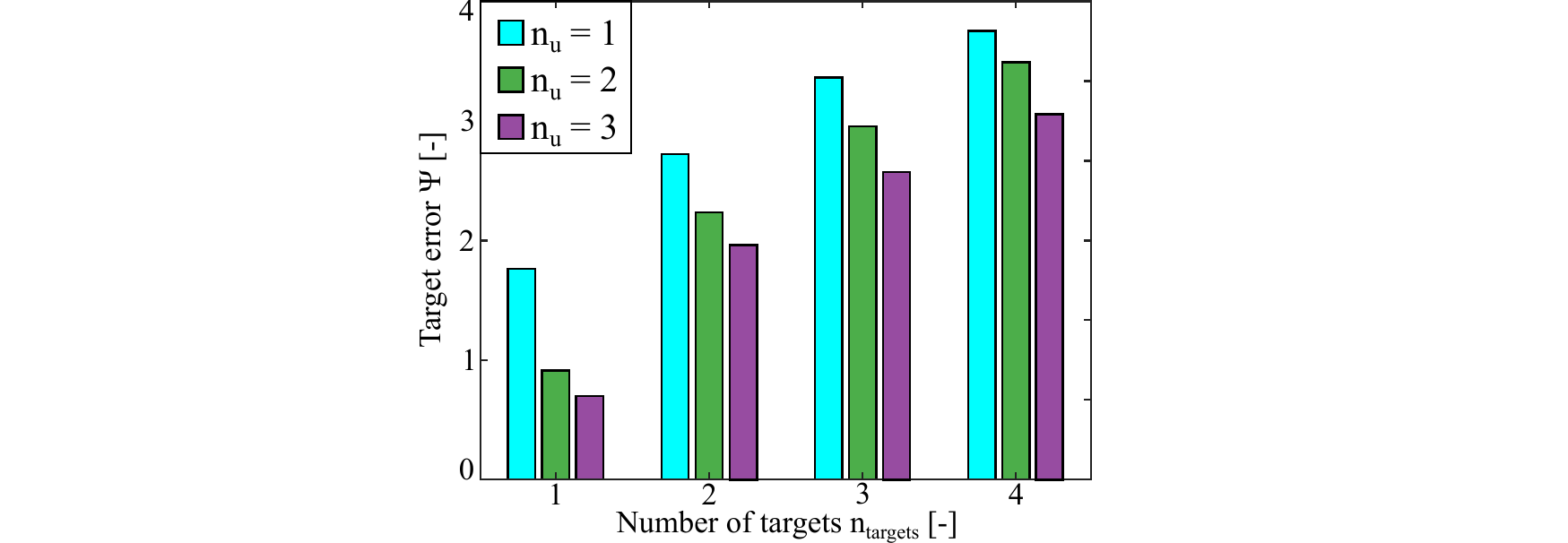}
\caption{\textbf{Random targets error.} For a random set of $n_{targets}$, we use the greedy algorithm to find the actuator design that minimizes the target error $\Psi$. Note that each target is bounded by a cubic domain centered with the actuator and norm equal to $1/2(n_u\cdot n_s)$ to ensure that it is within reach. We report the average error based $1000$ simulations with $n_{targets}\in\{1,2,3,4\}$ and $n_u\in\{1,2,3\}$. Note that for each set of targets, we fix $n_u$ and choose the $n_s$ (with the constraint that $n = n_u\cdot n_s \le 15$) that minimizes the error.}\label{1_design_fig6}
\end{figure}

\begin{figure}[H]
\centering
\includegraphics[scale=1]{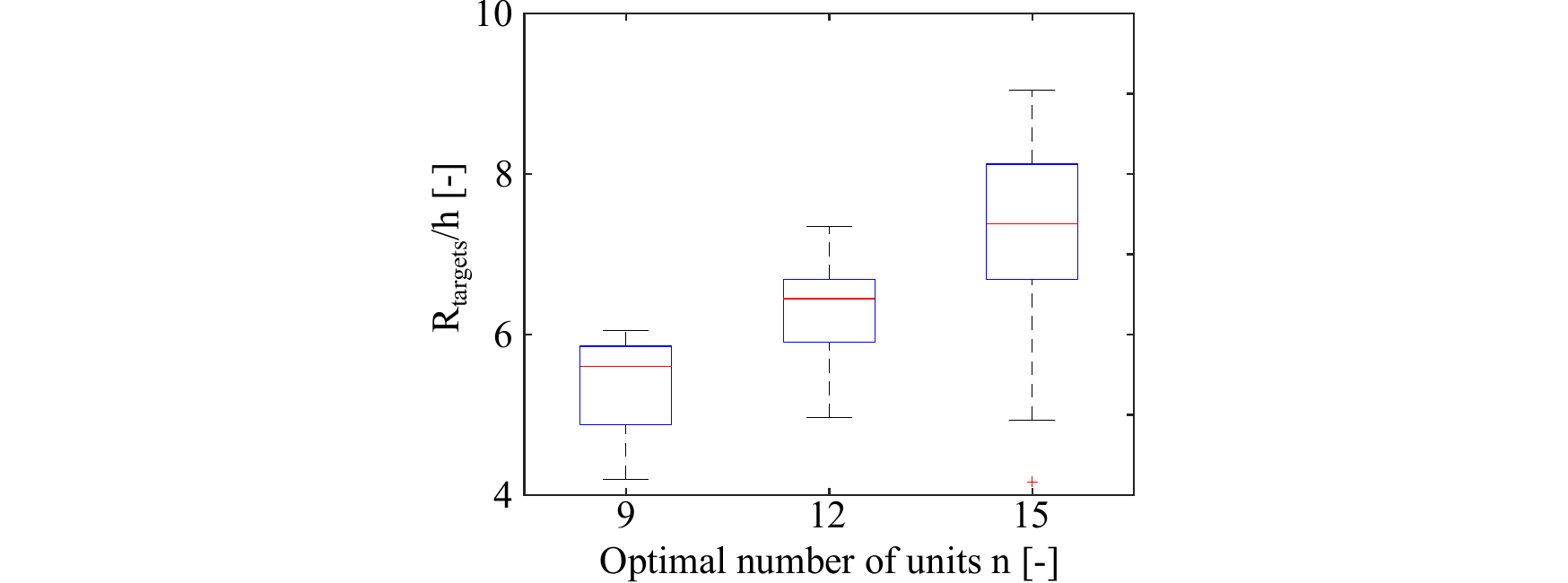}
\caption{\textbf{Optimal number of units as a function of the target radius.} For $100$ random sets of $n_{targets} = 3$, we measure the target radius, $R_{targets}$, i.e. the radius of the sphere fitted with the targets, and we use the greedy algorithm to find the number of units that minimizes the error between the target and the top cap's centroid. We report here the target radius normalized by the module height, $R_{targets}/h$, as a function of the optimal number of units $n$ found by the greedy algorithm. We find that the average target radius increases with the number of units to minimize the error $\Psi$.}\label{1_design_fig_radius}
\end{figure}

In Fig.~\ref{1_design_fig4} we present   results obtained when the greedy algorithm is programmed to maximize   $\theta_{act}$ (Fig.~\ref{1_design_fig4}a-c) and $||\mathbf{d}||/h$ (Fig.~\ref{1_design_fig4}d-f). Note that we present present results for three different values of $n_u$ (i.e. $n_u=1,$ 2 and 3).\\
\\
\noindent Then, in Fig.~\ref{1_design_fig5}  we consider a set of three targets $(T1,T2,T3)$ different from that included in the main text and present the results for the optimal design identified by the greedy algorithm. Next, in Fig.~\ref{1_design_fig7} we show the inverse design of an actuator reaching the same set of three targets considered in Fig.~\textcolor{blue}{3} of the manuscript, but with the additional constraint that the targets much be reached successively by decreasing pressure.\\
\\
\noindent Finally, in Fig.~\ref{1_design_fig6} we show how the minimum value of $\Psi$ found by the greedy algorithm varies with the number of targets, $n_{targets}$, and the units forming a super-cell, $n_u$, and in Fig.~\ref{1_design_fig_radius} how the target radius (i.e.~the radius of the sphere fitted with the targets) influences the optimal number of units of the actuator.

\end{document}